\documentclass[journal]{IEEEtran}
\ifCLASSINFOpdf
\else
\fi

\usepackage{algorithm}
\usepackage{algorithmic}
\usepackage{multirow}
\usepackage{latexsym}
\usepackage{subfigure}
\usepackage{graphicx}
\usepackage{amsmath}
\usepackage{amssymb}
\usepackage{subfigure}
\usepackage{subcaption}
\usepackage{tikz}
\usepackage{pgfplots}

\usepackage{makecell}
\usepackage{color}
\usepackage{gensymb}
\usepackage{bbding}
\usepackage{hyperref}
\hypersetup{hypertex=true,
	colorlinks=true,
	linkcolor=red,
	anchorcolor=red,
	citecolor=green}

\begin{document}
\title{MCSDNet: Mesoscale Convective System Detection Network via Multi-scale Spatiotemporal Information}

\author{Jiajun~Liang,
        Baoquan~Zhang, %~\IEEEmembership{Member,~IEEE,} %~\IEEEmembership{Member,~IEEE,}
	  Chuyao~Luo,   
        Xutao~Li, %~\IEEEmembership{Member,~IEEE,}
	    Yunming~Ye, %~\IEEEmembership{Member,~IEEE,}
        Xukai~Fu %~\IEEEmembership{Member,~IEEE,}
        %John~Doe,~\IEEEmembership{Fellow,~OSA,}
        %and~Jane~Doe,~\IEEEmembership{Life~Fellow,~IEEE}% <-this % stops a space
\thanks{Jiajun Liang, Baoquan Zhang, Chuyao Luo, Xutao Li, and Yunming Ye are with the School of Computer Science and Technology, Harbin Institute of Technology, Shenzhen, Shenzhen 518055, Guangdong, China; Xukai FU is with Beijing Jinkai New Energy Environmental Technology Co., Ltd.}% <-this % stops a space
\thanks{E-mail: liangjiajun2002@163.com, baoquanzhang@hit.edu.cn, \{lixutao, yeyunming\}@hit.edu.cn, luochuyao@stu.hit.edu.cn, 2510442827@qq.com}% <-this % stops a space
\thanks{Corresponding author: Baoquan Zhang.\protect\\}% <-this % stops a space
}


\markboth{Journal of \LaTeX\ Class Files,~Vol.~13, No.~9, September~2014}%
{Shell \MakeLowercase{\textit{et al.}}: Bare Demo of IEEEtran.cls for Journals}



\maketitle

\begin{abstract}
The accurate detection of Mesoscale Convective Systems (MCS) is crucial for meteorological monitoring due to their potential to cause significant destruction through severe weather phenomena such as hail, thunderstorms, and heavy rainfall. However, the existing methods for MCS detection mostly targets on single-frame detection, which just considers the static characteristics and ignores the temporal evolution in the life cycle of MCS. In this paper, we propose a novel encoder-decoder neural network for MCS detection(MCSDNet). MCSDNet has a simple architecture and is easy to expand. Different from the previous models, MCSDNet targets on multi-frames detection and leverages multi-scale spatiotemporal information for the detection of MCS regions in remote sensing imagery(RSI). As far as we know, it is the first work to utilize multi-scale spatiotemporal information to detect MCS regions. Firstly, we design a multi-scale spatiotemporal information module to extract multi-level semantic from different encoder levels, which makes our models can extract more detail spatiotemporal features. Secondly, a Spatiotemporal Mix Unit(STMU) is introduced to MCSDNet to capture both intra-frame features and inter-frame correlations, which is a scalable module and can be replaced by other spatiotemporal module, \emph{e.g.}, CNN, RNN, Transformer and our proposed Dual Spatiotemporal Attention(DSTA). This means that the future works about spatiotemporal modules can be easily integrated to our model. Finally, we present MCSRSI, the first publicly available dataset for multi-frames MCS detection based on visible channel images from the FY-4A satellite. We also conduct several experiments on MCSRSI and find that our proposed MCSDNet achieve the best performance on MCS detection task when comparing to other baseline methods. Particularly, MCSDNet shows remarkable capability in extreme conditions where MCS regions are densely distributed. We hope that the combination of our open-access dataset and promising results will encourage the future research for MCS detection task and provide a robust framework for related tasks in atmospheric science. Our code is available \footnote {\url{https://github.com/250HandsomeLiang/MCSDNet.git}}.
\end{abstract}


% \fi
\IEEEpeerreviewmaketitle

\section{Introduction}
\label{section1}
\IEEEPARstart{c}{onvective} weather is a type of sudden and intense meteorological disaster that can cause significant destruction. It is often accompanied by severe weather phenomena such as hail, thunderstorms, strong winds, and short-term heavy rainfall, posing a serious threat to the life and property safety of the public. Cloud masses generated by convection are commonly referred to as convective clouds. The air inside these clouds is in an extremely unstable state, characterized by high temperatures in the lower atmosphere and low temperatures in the upper atmosphere. Convective weather, with its short occurrence time, sudden intensity, and significant impact, has always been a challenging aspect in meteorological monitoring. 
\pgfplotstableread[row sep=\\,col sep=&]{
    interval & value\\
    0\%-1\%  & 69.27  \\
    1\%-2\%  & 15.61 \\
    2\%-3\%  & 9.43 \\
    3\%-4\%  & 4.18\\
    4\%-5\%  & 1.51   \\
    5\%-     & 0 \\
}\mydata
\begin{figure}
	\centering
        \begin{tikzpicture}
            \begin{axis}[
                    ybar,    
                    symbolic x coords={0\%-1\%, 1\%-2\%,2\%-3\%,3\%-4\%,4\%-5\%,5\%-},
                    xtick=data,
                    nodes near coords={\pgfmathprintnumber{\pgfplotspointmeta}\%},% 设置标签显示格式为百分比
                    % nodes near coords,
                    width=\columnwidth,
                    ymin=0,ymax=100,
                    xlabel={Percentage},
                    ylabel={MCS region cover proportion},
                    bar width=20pt % 设置柱子的宽度为10单位
                ]
                \addplot [draw=none,fill=blue!50!cyan] table[x=interval,y=value]{\mydata};
            \end{axis}
        \end{tikzpicture}
	\caption{The proportion distribution of MCS regions among the observation data collected in China in 2018.}
	\label{fig5}
\end{figure}
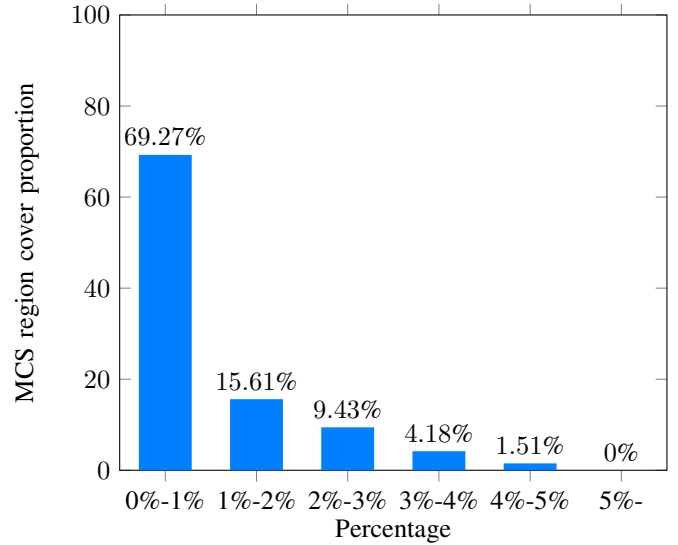
Convective weather is caused by mesoscale convective system(MCS), which is a cloud cluster with a convective core and extends approximately 100 km in a direction, forming a general precipitation area \cite{fiolleau2013algorithm}. The key of convective weather detection is how to detect MCS regions. With the advancement of remote sensing satellite technology, researchers have attempted to utilize remote sensing imagery(RSI) to detect MCS \cite{wang2023convection} \cite{yang2023convective}. However, MCS detection faces the issue of class imbalance, where the number of negative samples far exceeds the number of positive samples. As shown in Fig. \ref{fig5}, among the observation data collected in China in 2018, more than 60\% of satellite images contained only 1\% of MCS regions. Additionally, MCS has a complex and dynamic life cycle, which consists of three stages: initiation, maturity, and a decaying phase where the location and shape of convective cloud change over time. When performing detection of MCS regions, it is necessary to consider both the spatial and temporal information. MCS detection is a challenging issue with few existing solutions but is worth to be addressed.

In terms of MCS detection, fixed brightness temperature threshold method(BT) is the most commonly used detection technique. It distinguishes MCS regions from other objects at the pixel level by utilizing their spectral and wavelength differences \cite{zhu2012object}. However, the BT threshold is not universally and it can produce significant detection differences due to varying conditions such as geographical location, temperature, and properties of the lower atmosphere. 

In recent years, researchers have tried to apply machine learning model to atmospheric science \cite{zuo2022identification} \cite{utsav2017statistical}. Different from BT threshold, the machine learning model doesn't need to understand complex atmospheric physical and dynamic processes and it is a data-driving method. However, MCS detection algorithms based on machine learning are not end-to-end solutions. Instead, their detection effectiveness heavily relies on the quality of feature engineering, resulting in insufficient intelligent decision-making capabilities \cite{le2017deep}. 

Benefiting from the development of deep learning, end-to-end models based on neural network have been applied to RSI analysis, \emph{e.g.}, hyperspectral image classification \cite{li2019deep} \cite{10153685} \cite{10497695}, cloud detection \cite{basaeed2016supervised} \cite{mohajerani2019cloud} \cite{yang2019cdnet}, precipitation nowcasting \cite{10309842} \cite{10403855} and so on. However, these models are not designed for MCS detection and they don't take into account the dramatic changes of MCS in the whole life cycle. Besides, the existing models target on single-frame image and don't consider the temporal evolution in the life cycle of MCS. It is challenging to detect MCS regions accurately.

Similar to \cite{9686686} \cite{10497698}, MCS detection task can be regarded as a semantic segmentation task in computer vision. Given a satellite image, we aim to use a specific algorithm to classify each pixel in the original image . Since each pixel in the original image can only belong to either the category of MCS region or not, the task of MCS detection can also be considered as a pixel-level binary classification task. We can utilize the efficient models in semantic segmentation task to detect MCS regions.

% The fully convolutional network(FCN) \cite{long2015fully} replaces the linear layer with convolution layer, accepts arbitrary large images as input and predicts the class label of each pixel. Though the models based on FCN architecture is simple, it has gained significant improvement in cloud detection \cite{jeppesen2019cloud} \cite{dronner2018fast}.
These methods mentioned above all target on single-frame detection and ignore the temporal evolution in the life cycle of MCS. Considering the dramatic and changeable characteristics of MCS, they are limited to achieve excellent performance. In this paper, we argue the significance of spatiotemporal information and propose an encoder-decoder neural network for MCS detection(MCSDNet). Different from the existing models, MCSDNet targets on multi-frames MCS detection and takes image sequence as input. This model can capture the temporal evolution of MCS by exploring cross-frames variations in image sequence. In MCSDNet, we capture multi-scale spatiotemporal information by incorporating the feature maps from different encoder levels. Then, we apply a Spatialtemporal Mix Unit(STMU) to explore the relation and consistency between consecutive frames. STMU is a scalable spatiotemporal module to capture temporal evolution of MCS. In our implementation, we design a new spatiotemporal module named Dual Spatiotemporal Attention(DSTA) as STMU. Different from previous spatiotemporal modules \cite{shi2015convolutional} \cite{dosovitskiy2020image} \cite{tan2023temporal}, DSTA can effectively capture both intra-frame features and inter-frame correlations. Compared to other models, MCSDNet considers both spatial and temporal information of MCS regions and gains better performance in MCS detection task. 

Our main contributions can be summarized as follows:
% 使用定标后的可见光通道
% 创新点:1.提出基于时空数据的语义分割模型，能同时结合时间信息和空间信息进行对流云检测，提高了洁厕的精度，同时与以往只考虑单一尺度的模型不同，我们的模型引入了多尺度的时空信息，更加的适合对流云分布密集的极端情况。 2.MCSDNet是一种简单可复用的模型结构，拓展性强。MCSDNet通过STMU提取时空信息，STMU可以替换成任意的时空模块，例如卷积，RNN，transformer。在实验中，我们提出了一种新的时空融合模块Dual Spatiotemporal Attention Module,与之前的研究相比，我们同时聚焦于时间信息和空间信息，而不是单一的时间或空间维度。取得了更好的性能。 3.设计了一个新的数据集，用于基于时空数据的对流云检测。
\begin{itemize}
	\item We propose an encoder-decoder neural network for MCS detection(MCSDNet). MCSDNet has a simple architecture and is easy to expand. The model utilizes a Spatiotemporal Mix Unit(STMU) to capture temporal evolution of MCS regions, which is scalable modules and can be replaced by other spatiotemporal modules, \emph{e.g.}, CNN, RNN, Transformer and our proposed Dual Spatiotemporal Attention. Besides, MCSDNet maintains the original performance standards of the spatiotemporal modules without any compromise. This means that the future works about spatiotemporal modules can be easily incorporated into our model.
	\item Different from the existing methods, our model targets on multi-frames MCS detection and captures the temporal evolution in the life cycle of MCS by exploring the cross-frames correlations among image sequence. In MCSDNet, we design a multi-scale spatiotemporal information module to extract multi-scale semantic from different encoder levels, which makes out model more suitable to extreme conditions. As far as we know, it is the first work to utilize multi-scale spatiotemporal information to detect MCS regions. 
	\item We create and publicly share a Mesoscale Convective System Detection dataset based on the captured images from FY-4A satellite: Mesoscale Convective System Remote Sensing Image(MCSRSI). As far as we know, it is the first publicly available dataset for multi-frames MCS detection task based on  visible channel images. Besides, we conduct experiments on it. The results demonstrate that MCSDNet can achieve state-of-the-art performance on MCS detection task. It provides a baseline for future research in this direction.
\end{itemize}

The rest of this work is organized as follows: In Section~\ref{section_2}, we briefly review related works on MCS detection. Section~\ref{section_3} describes our model(MCSDNet) in details. In Section~\ref{section_4}, we introduce the newly created dataset for MCS detection: MCSRSI and present the experimental results on it. Finally, the conclusion is summarized in Section~\ref{section_5}.
\section{Related Work}
\begin{figure*}
	\centering
        % \flushright
	\includegraphics[width=.85\textwidth]{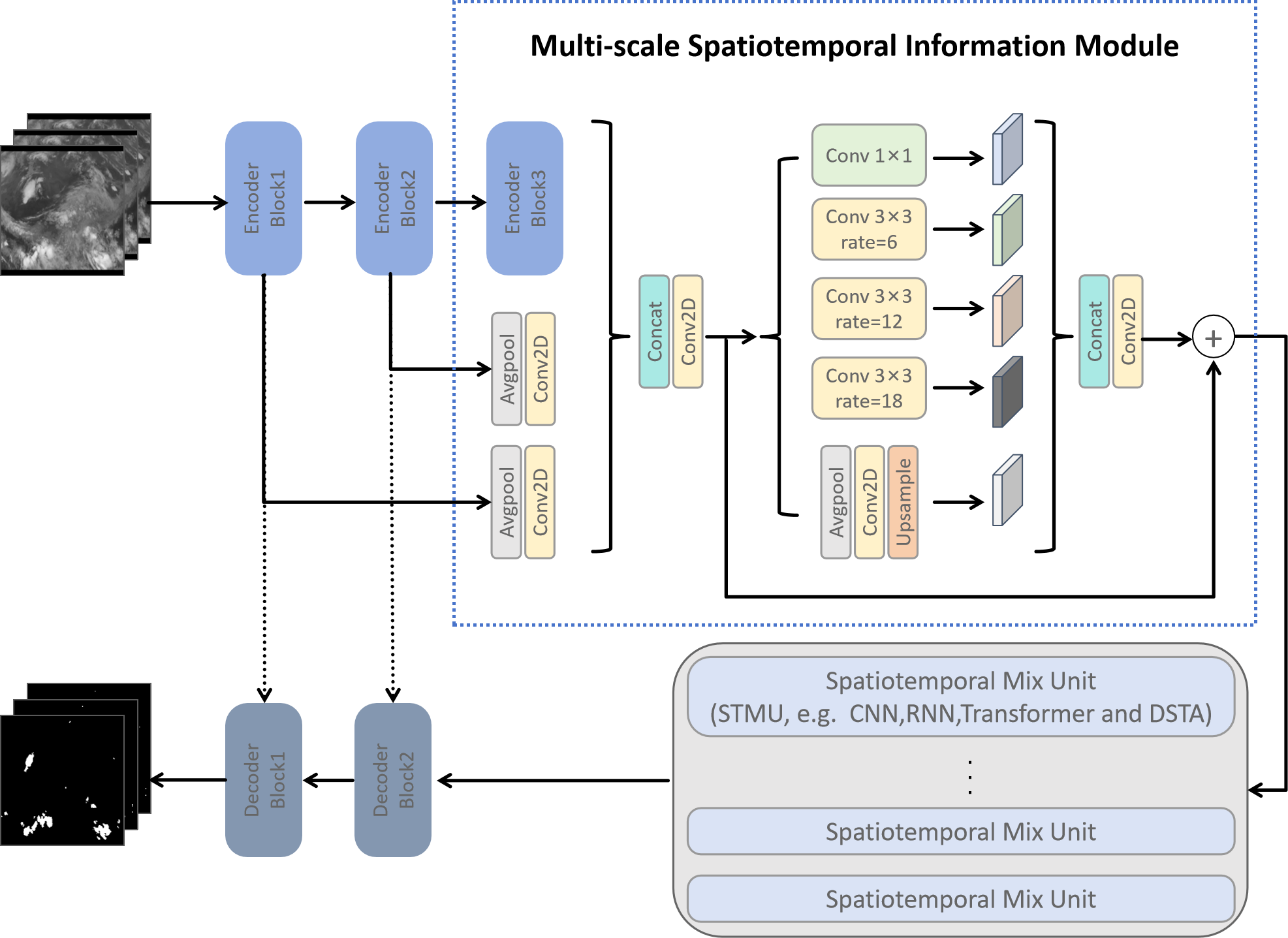}
	\caption{Architecture of the proposed MCSDNet. A sequence of images is processed in parallel by a shared convolutional encoder. We introduce multi-scale spatiotemporal information by incorporating the feature maps from different encoder levels. At the lowest resolution, several Spatiotemporal Mix Units(STMUs) explore the spatial and temporal dependencies between the different frames. In the decoder, semantic information is transferred from the encoder to decoder to enhance the detection results further.
    }
	\label{fig1}
\end{figure*}
\label{section_2}
%-------------------------------------------------------------------------
% 1.threshold methods 2. machine learning 3. deep learning
\subsection{Mesoscale Convective System Detection Methods}
When detecting MCS regions, threshold methods is the most commonly used methods \cite{schumacher2005organization}, \cite{haberlie2019radar}, \cite{chen2019mesoscale}, \cite{huang2018long}. Brightness temperature(BT) method utilizes the relatively low cloud top temperature in convective clusters to detect MCS. The threshold is usually ranging from 208 K to 255 K \cite{machado1992structural} \cite{machado1998life}. BT difference(BTD) is based on water vapor sensitivity and uses the brightness temperature difference to explore convective clouds. Different from threshold methods, machine learning methods do not need to understand complex atmospheric physical and spectral channel information. It is gradually applied to MCS detection. Kim \textit{et al.} \cite{kim2017detection} utilize the images from the Fengyun-8 satellite to construct random forest and logistic regression models which can effectively locate MCS regions. Yang \textit{et al.} \cite{yang2023convective} construct MCS detection and tracking algorithms for the Himawari-8 advanced Himawari-8 imager (AHI) data by combining machine learning models, area overlapping, and Kalman filter algorithms. 

In recent year, deep learning has gained great breakthrough in computer vision tasks \cite{lecun2015deep}. Some researchers attempt to utilize deep learning model to detect MCS region. CloudFU-Net \cite{10500381} is a fine-grained segmentation method for ground-based cloud images which is based on encoder-decoder structure. Compared to threshold methods and machine learning methods, it possess better generalization performance and efficiently segment different cloud genera. Convection-Unet \cite{wang2023convection} has a simple architecture and utilizes the Unet architecture for pixel-level convection detection. FTransUNet \cite{ma2024multilevel} incorporates shallow-level and deep-level features in a multilevel manner, which apply multi-modal information on RSI semantic segmentation. However, these methods mentioned above are designed for single-frame MCS detection. They ignore the dramatic change and complex development of MCS. There is no special design to extract temporal evolution in the life cycle of MCS.
\subsection{Semantic Segmentation}
The semantic segmentation methods are designed to classify
each pixel of an image and assign a semantic tag to each
pixel. Unet \cite{ronneberger2015u} is a u-shaped architecture network. It consists of a contracting path and an expansive path. The contracting path downsamples the images and extracts features from them. The expansive path upsamples the images and classifies each pixel. Deeplabv3+ \cite{chen2018encoder} is based on encoder-decoder architecture. It utilizes atrous convolution and atrous spatial pyramid pooling(ASPP) to extract multi-scale semantic information. In Swin Transformer \cite{liu2021swin}, a pyramid-like architecture is adopted to address the multi-scale characteristics of objects. It uses Transformer for dense predicting.  However, the semantic segmentation models are not designed for remote-sensing images, and they are not robust enough for MCS detection. These models take a single frame of image sequence as input and ignore cross-frames consistent features. Due to the development of MCS exhibits strong temporal patterns, we need to consider both spatial and temporal information when detecting MCS regions.

% In recent years, some researchers integrate semantic segmentation task to video domain. Video semantic segmentation focus on classifying each pixel in consecutive video frames into special category. 
\subsection{Video Understanding}
Video Understanding task aims to analyze each frames in video and then classify the video or generate the future frames. Accurate video understanding yield substantial benefits across a wide range of practical applications, including climate change analysis \cite{reichstein2019deep} \cite{shi2015convolutional}, forecasting human motion \cite{wang2018rgb}, predicting traffic flow \cite{fang2019gstnet}, and representation learning \cite{jenni2020video}. Inspired by the success of long short-term memory (LSTM) \cite{hochreiter1997long} in sequential modeling, ConvLSTM \cite{shi2015convolutional} leverages convolutional neural networks to model the LSTM architecture and represents a pioneering contribution to the field of video understanding. PredRNN \cite{wang2017predrnn} introduces a spatiotemporal LSTM unit which can effectively extract and retains spatial and temporal representations concurrently. Its subsequent iteration, PredRNN++ \cite{wang2018predrnn++}, enhances this by incorporating gradient highway units and casual LSTM, enabling adaptive capture of temporal dependencies. With the development of vision transformer architecture, \cite{bertasius2021space} \cite{arnab2021vivit} \cite{liu2022video} apply different space-time attention strategies to extract multi-scale spatiotemporal information and achieve great performance in video understanding task. In \cite{tan2022simvp}, Tan \textit{et al.} introduce SimVP, a simple architecture which is completely built upon convolutional networks without recurrent architectures. Without introducing any extra tricks and strategies, SimVP can achieve state-of-the-art performance on various benchmark datasets. Inspired by the breakthrough in video understanding, MCSDNet utilizes both spatial and temporal information to detect MCS regions. 

\subsection{Vision Transformer}
Transformer is a simple network architecture based solely on attention mechanisms, dispensing with recurrence and convolutions entirely \cite{vaswani2017attention}. Transformer is first proposed in natural language processing(NLP) tasks and gains significant improvements in various downstream tasks, \emph{e.g.}, machine translation, name entity recognition and large language model. Motivated by the remarkable performance achievements of Transformer in natural language processing (NLP), researchers have tried to explore the application of Transformer to computer vision tasks and started the investigation on ViTs. ViT proposes a pure transformer which is applied directly to sequence of image patches and performs very well on image classification tasks. Swin Transformer \cite{liu2021swin} designs a pyramid-like architecture and utilizes shifted windowing scheme to improve the efficiency of self-attention computation. Although ViTs have gained a great performance in natural image analysis, they are not well-suited for satellite image analysis. 

\section{Methodology}
\begin{figure}
	\centering
	\includegraphics[width=.95\columnwidth]{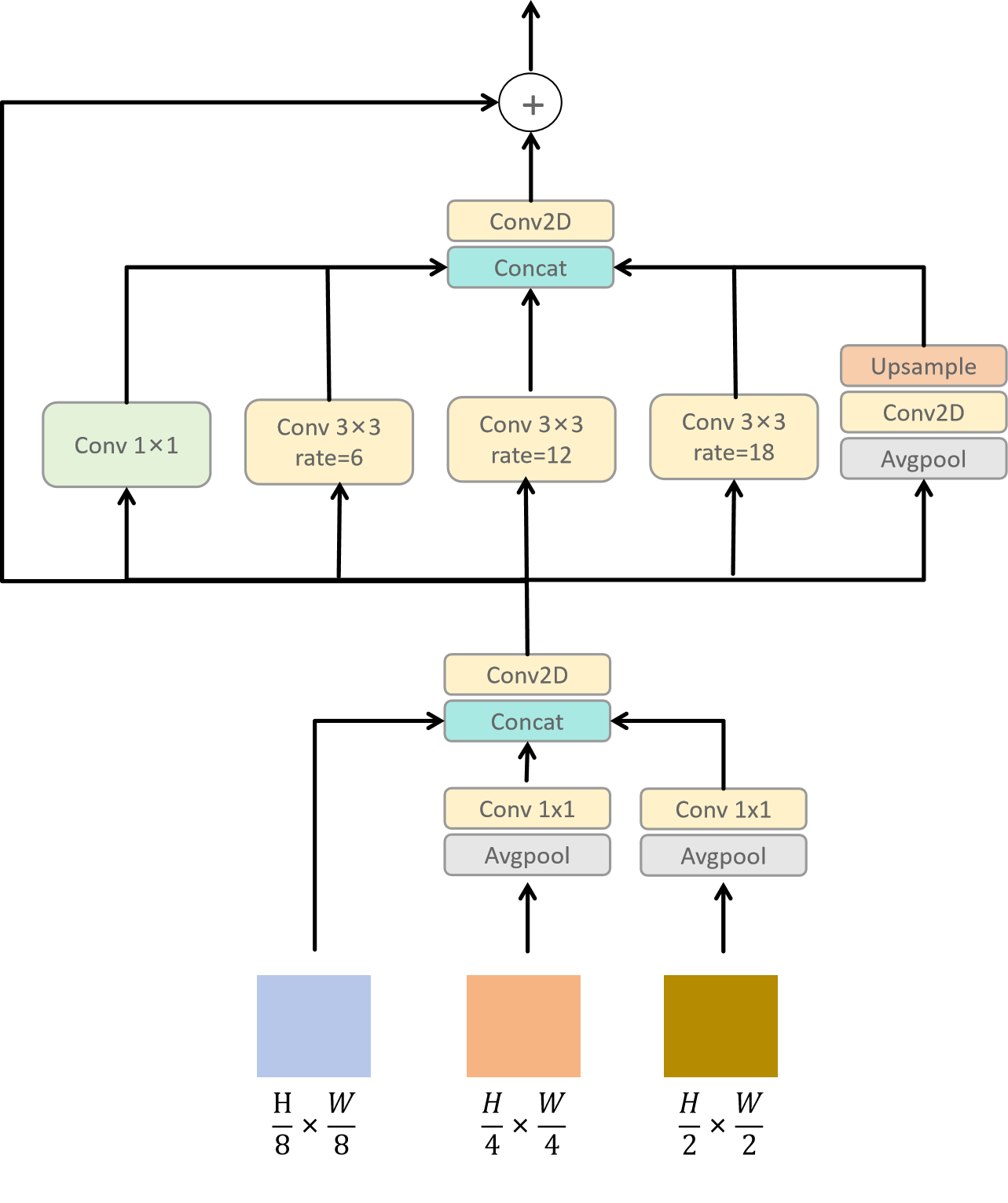}
	\caption{Multi-scale spatiotemporal information module.}
	\label{fig22}
\end{figure}
% 介绍结构
\label{section_3}

%-------------------------------------------------------------------------
\subsection{Preliminaries}
\label{section_3_1}
We consider the MCS detection task as RSI sequence segmentation task. Given a RSI sequence $X\in R^{T \times C \times H \times  W}$ ,  with T
the length of the sequence, C the number of channels, and H × W the spatial extent, we aim to predict the segmentation result  $Y\in R^{T \times C \times H \times  W}$  of these images.

The model with learnable parameters $\theta$ learns a mapping $F_{\theta}:X^{T \times C \times H \times W } \longmapsto Y^{T \times C \times H \times W } $ by exploring both spatial and temporal dependencies. In our case, the mapping $F_{\theta}$ is a fully convolutional network(FCN)\cite{long2015fully} trained to minimize the difference between the predicted segmentation  result and the ground-truth segmentation result. The optimal parameters $\theta^{*}$ are
\begin{equation}
	\begin{aligned}
                \theta^{*} = arg \min_{\theta} L(F_\theta(X),Y),
	\end{aligned}
	\label{eq1}
\end{equation}
where L is a loss function that evaluates such differences. We adopt the FocalLoss as the loss function to calculate the loss between the predict result and the ground truth
\begin{equation}
	\begin{aligned}
                L_{FocalLoss} = -\sum{(1-p_t)^{\gamma}log(p_t)},
	\end{aligned}
	\label{eq2}
\end{equation}
where $p_t$ reflects the degree of closeness to the ground truth and $\gamma $ is adjustable factor. A larger $p_t$ indicates a closer proximity to ground truth.
\subsection{Overall Framework}
\begin{figure*}
	\centering
	\includegraphics[width=\textwidth]{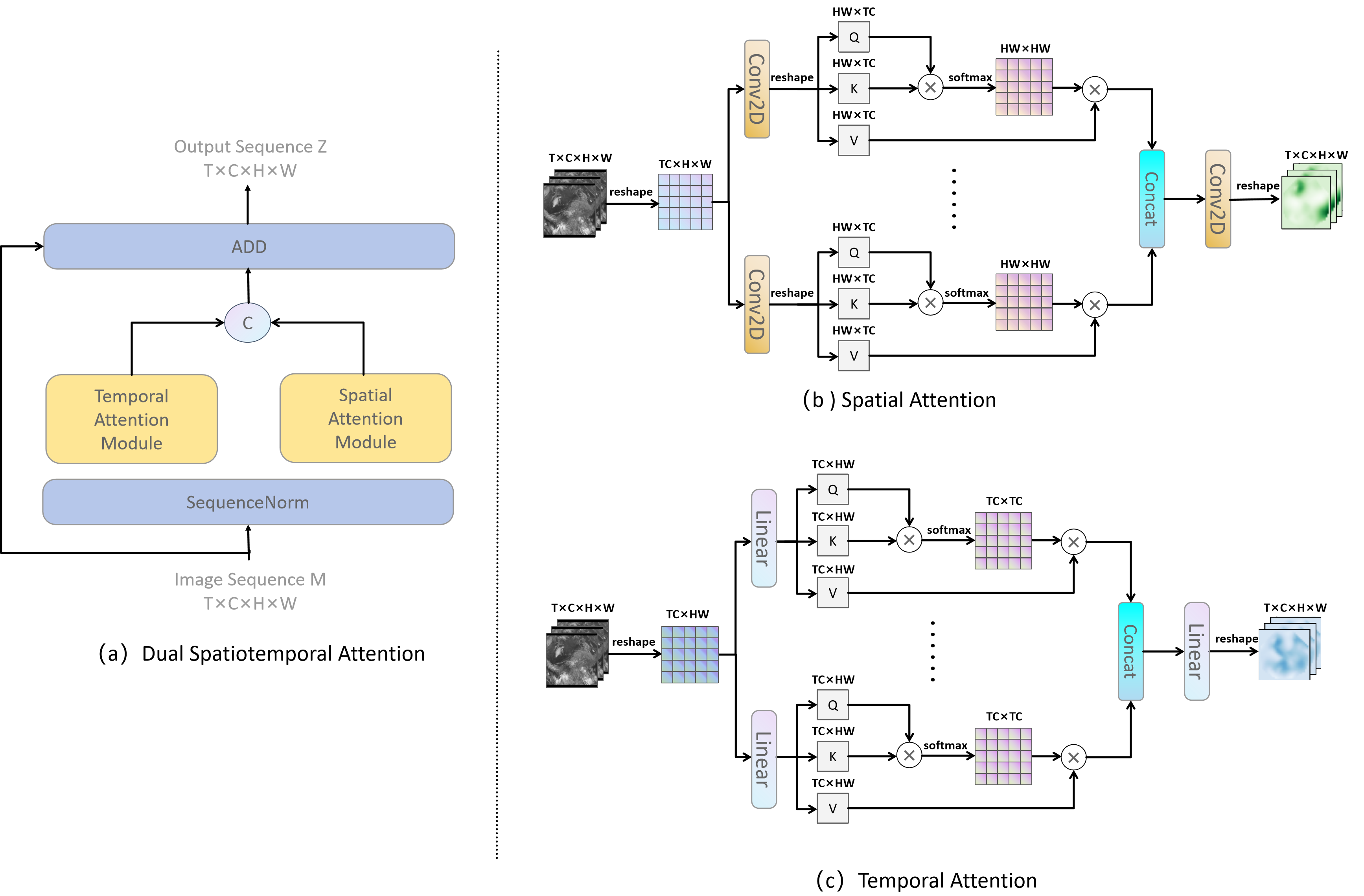}
        \caption{The detail of proposed Dual Spatiotemporal Attention. (a) Architecture; (b) Spatial Attention Module; (c) Temporal Attention Module.}
	\label{fig2}
\end{figure*}
\label{section3_2}
% \begin{figure}
% 	\centering
% 	\includegraphics[width=\columnwidth]{image/Texure.png}
% 	\caption{Proposed Texture Module, which extracts multi-scale spatial information and enhances the texture feature of convective cloud. The dilation rates equal to 1,6,12,18. 
%         }
% 	\label{fig2_2}
% \end{figure}

Fig. \ref{fig1} shows the architecture of the proposed MCSDNet. MCSDNet is based on an encoder–decoder architecture, which consists of three parts: an Encoder responsible for extracting spatial information, Spatiotemporal Mix Units (STMUs) responsible for modeling spatial and temporal information in image sequence, and a Decoder responsible for upsampling and reconstructing features from low-resolution feature maps.

In the encoder, MCSDNet adopts a RSI sequence $Z\in R^{T \times C \times H \times  W}$ as input. Then, the model extracts the spatial features by several ConvNormReLU blocks(Conv2d+GroupNorm+ReLU). To reduce computational complexity, the different frames in sequence share the same multi-level spatial convolutional encoder. Each layer in encoder is formulated as
\begin{equation}
	\begin{aligned}
                Z_{i}=\sigma(Norm2d(Conv2D(Z_{i-1}))),
	\end{aligned}
	\label{eq3}
\end{equation}
where $\sigma$ is a non-linear activation function and Norm2d is a normalization layer. 
To maintain diversity between images at different time frames, we use Group Normalization with 4 groups instead of Batch Normalization in the encoder. Besides, each layer in encoder begins with MaxPool with 2 scale factor to downsample the output of the front layer. 

STMU is a scalable module, which can be replaced by other spatiotemporal modules. STMU takes the multi-scale spatiotemporal information from encoder as input and then explores the spatial and temporal dependencies between the different frames, which embeds the global context spatiotemporal information into image at different time. The development process of MCS exhibits strong temporal patterns. It can significantly improve the accuracy of MCS detection by introducing temporal information. In our implementation, we design a new module named Dual Spatiotemporal Attention(DSTA) as STMU. Different from the previous modules, DSTA can capture both intra-frame features and inter-frame correlations.

In the decoder, the model reconstructs the feature maps gradually. The lost spatial
information in downsampling is addressed in the extremely lightweight decoder by using a skip connection from the encoder to the decoder. The detection accuracy is improved with a lower complexity increase \cite{luo2023trcdnet}. Each layer in decoder is formulated as
\begin{equation}
	\begin{aligned}
                Z_{i}=\sigma(Norm2d(Conv2D([e_i|unConv2D(Z_{i-1})]))),
	\end{aligned}
	\label{eq4}
\end{equation}
where unConv2D is an upsmaple layer, $e_i$ is the output of the $i^{th}$ layer of the encoder, and $[ | ]$ is the concatenation operation.

\subsection{Multi-scale Spatiotemporal Information Module}
MCS is characterized by dense local distribution and sparse global distribution. We attempt to introduce multi-scale spatiotemporal information to improve the performance of MCS detection. As shown in Fig. \ref{fig22}, we design a multi-scale spatiotemporal information module to incorporate multi-level semantic of feature maps. In MCSDNet, we first apply global average pooling on high resolution feature maps from encoder and feed them into a 1$\times$1 convolution to resize them into the same shape as the last feature map of encoder. With the global average pooling, we can reduce the loss of information during the process of changing the resolution of feature maps. Then, we concatenate the output from different encoder levels and pass them through a 1$\times$1 convolution to generate a feature map with multi-scale spatiotemporal information. 

The distribution of MCS is influenced by season and geographical location, making it more challenging to segment objects than that of natural images. The feature maps in different conditions have different optimal field-of-view. When detecting MCS regions, we desire more detailed spatiotemporal information. According to \cite{chen2017rethinking}, we design a pyramid pooling module to extract multi-scale context information. With the pyramid pooling module, our model can adaptively adjust the field-of-view when keeping the resolution of feature maps, which captures MCS regions at multiple scales. The pyramid pooling module consists of several parallel atrous convolutions with different atrous rates and a global average pooling. We apply atrous convolutions to extract multi-scale information from RSIs which makes our model can capture MCS features from multiple field-of-views. Besides, the global average pooling incorporates global context information to the model and adopts the image level features. Then, we concatenate the results from all branches and feed them into a 1$\times$1 convolutions to generate the final feature map.   
% According to \cite{chen2018encoder}, we designed a Texture Module (see in Fig. \ref{fig2_2}) to capture multi-scale spatial information of MCS. Different from natural image, remote sensing image exhibits imbalance of MCS region and non-MCS region, and they have similar feature expression. To distinguish them, our proposed Texture Module extracts multi-scale context information via astrous convolution and enhances the texture feature expressions of convective cloud with surrounding cloud pixels. 

% The Texture Module receives the last output of encoder, and then extracts the long-range texture information by atrous convolution, which is significant to locate MCS regions. After that, the feature map from Texture Module is passed to STMU and it will explore the cross-frame relationship. By utilizing atrous convolution to capture features, the Texture Module can increase the receptive field without losing spatial information, thereby obtaining more semantic information and improving the accuracy of MCS detection. 
\subsection{Spatiotemporal Mix Unit}
\begin{figure}
	\centering
	\includegraphics[width=.5\columnwidth]{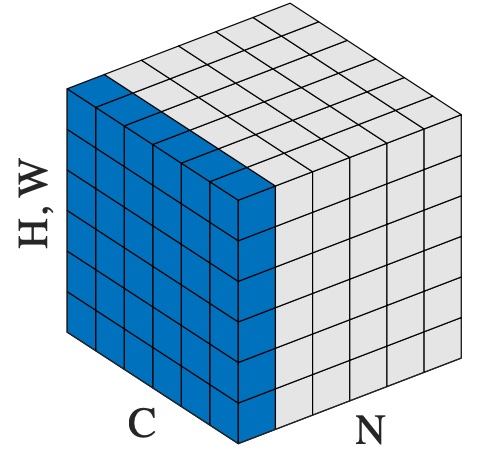}
	\caption{Sequence Normalization which normalize the batch data along 
        $T \times C \times H \times W$ dimension. T and C are concatenated together during normalization. }
	\label{fig3}
\end{figure}
MCSDNet applies Spatiotemporal Mix Unit(STMU) to capture the temporal evolution of MCS life cycle, which is important to detect MCS regions. Our proposed model has a simple architecture and is easy to expand. It can choose different STMU for sequence modeling, \emph{e.g.}, CNN, RNN and Transformer. To effectively utilize the spatiotemporal features of MCS regions, we proposed a new STMU named Dual Spatiotemporal Attention(DSTA). Compared to the previous works \cite{wang2021temporal} \cite{tan2023temporal}, DSTA simultaneously captures intra-frame features and inter-frame correlations. The previous STMU only take account into temporal evolution of MCS regions and ignores the texture features of convective cloud clusters which are important indicators to distinguish MCS regions and non MCS regions. In Section \ref{section_4}, we conduct an experiment to explore the performance of different STMU.

As shown in Fig. \ref{fig2}, DSTA consists of a temporal attention module(T-MSA) and a spatial attention module(S-MSA), which enable DSTA to extract both spatial features and temporal dependencies from RSI sequence. T-MSA performs self-attention operations along the temporal dimension, enabling it to model relationships across frames and capture the temporal evolution of MCS regions. Similarly, S-MSA performs self-attention operations along the spatial dimension, enabling it to capture long-range contextual semantic information and enhance the texture features of MCS regions. In addition, to maintain intra-sample diversity, we apply Sequence Normalization Fig. \ref{fig3} to normalize the batch data, the normalization dimension is $T \times C \times H \times W$ , where T and C are concatenated together during normalization. Sequence Normalization is formulated as 
\begin{equation}
	\begin{aligned}
                y=\frac{x-E[x]}{\sqrt{Var[x]+\varepsilon }}*\gamma+\beta,
	\end{aligned}
	\label{eq5}
\end{equation}
where E[x] and Var[x] are mean and standard-deviation calculated over $T \times C \times H \times W$ dimension and $\gamma$ and $\beta$ are learnable parameters. By introducing DSTA, our model can not only consider the texture features of MCS regions at the current moment but also adopt the motion distribution of MCS regions in previous and subsequent moments.  

DSTA adopts RSI sequence $X\in R^{T \times C \times H \times W}$ as input to explore the spatial and temporal dependencies between different frames and embeds the global context semantic into each frame. First, DSTA normalizes the RSI sequence along $T \times C \times H \times W$ to maintain the intra-sample diversity and accelerate the convergence speed of the model. Then, the parallel attention modules perform self-attention operations along TC dimension and HW dimension to capture intra-frame features and inter-frame correlations. Besides, we concatenate the outputs from two attention modules and feed them into a convolution layer to obtain better feature representation. To avoid performance degradation caused by gradients vanishing, the feature map after aggregation is reshaped to ${T \times C \times H \times W}$ and skip-connected to the input RSI sequence. The above process can be formulated as : 
\begin{equation}
	\begin{aligned}
	X^i=X^{i-1}+[T-MSA(SequenceNorm(X^{i-1})) \\
                ,S-MSA(SequenceNorm(X^{i-1}))], 
	\end{aligned}
	\label{eq6}
\end{equation}
where $X^i$ is the output from DSTA, T-MSA is the temporal attention module, S-MSA is the spatial attention module and [ ] is the concatenate operation. With the relationship modeling capability provided by multi-head self attention, DSTA can efficiently embed global context semantic information into each frame in RSI sequence, leading to detect MCS regions more accurately.

\section{Performance Evaluation}
\label{section_4}
%In this section, we evaluate the performance of MetaNODE on both transductive and inductive FSL setting, respectively.

\subsection{Datasets and Settings}

\begin{table}
        \renewcommand\arraystretch{1.5}
	\caption{The distribution of images in MCSRSI.}
	\begin{center}
		\smallskip\scalebox
		{1.0}{
		      \begin{tabular}{l c c c}
				\hline 
                Season & Time & Train & Test \\ 
                \hline \hline 
                \multirow{3}{*}{Spring} & 2018-03 & 425 & 106 \\ 
                                        & 2018-04 & 492 & 122 \\
                                        & 2018-05 & 588 & 146 \\ 
                \hline 
                \multirow{3}{*}{Summer} & 2018-06 & 516 & 129 \\ 
                                        & 2018-07 & 520 & 130 \\
                                        & 2018-08 & 627 & 156 \\ 
                \hline 
                \multirow{3}{*}{Autumn} & 2018-09 & 525 & 131 \\ 
                                        & 2018-10 & 536 & 133 \\
                                        & 2018-11 & 632 & 157 \\ 
                \hline 
                \multirow{3}{*}{Winter} & 2018-12 & 528 & 131 \\ 
                                        & 2018-01 & 520 & 129 \\
                                        & 2018-02 & 630 & 157 \\ 
                \hline 
                Summary & \#\#\# & 6539 & 1627 \\
                \hline
		\end{tabular}}
	\end{center}
	\label{table1}
\end{table}
To evaluate the proposed MCSDNet, we conduct a comprehensive evaluation by comparing it with other RSI analysis, semantic segmentation and video understanding methods. These evaluations are performed on the newly introduced remote-sensing image datasets named MCSRSI. Besides, we design a sub-dataset of MCSRSI to explore the performance of the MCSDNet under extreme condition, \emph{e.g.}, dense distribution of MCS regions. Then, we conduct ablation studies. The models are compared from both performance and efficiency aspects.
~\\
~\\
\noindent \textbf{Datasets}
Due to the absence of publicly available MCS detection datasets derived from geostationary satellites, to evaluate the performance of MCSDNet, we create a dataset based on Fengyun-4A(FY-4A) for MCS detection task named MCSRSI. FY-4A is the first satellite of the FengYun-4 Geostationary meteorological satellite. The radiometric imaging channels of FY-4A increase from 5 on the Fengyun-2G(FY-2G) satellite to 14, covering visible light, shortwave infrared, mid-wave infrared, and long-wave infrared bands. It can capture the cloud map of the Earth every 15 min. MCSRSI is the first large-scale, publicly available dataset for MCS detection task based on visible channel. This dataset, as well as more information about its composition, are publicly available at \url{https://github.com/250HandsomeLiang/MCSRSI.git}.
  
MCSRSI is comprised of 8157 images in China region from FY-4A of shape $800 \times 1280$, ranging from January 2018 to December 2018 (see Table. \ref{table1} ). The time between acquisitions is 15 minutes. After calibrate the 12th channel of FY-4A’s L1 data, we acquire visible channel image as model's input. With the help of domain experts, we utilize brightness temperature threshold and radar echo intensity to manually label MCS regions. MCS exhibits significant differences in brightness temperature in different regions. In the southern of China, MCS regions generally have lower brightness temperature threshold, ranging from 215K to 225K. However, in the northern of China, they tend to range from 220K to 235K, and even reach around 240K in some cases. In terms of radar echo intensity, convective cloud typically has radar echo intensity greater than 25dbz. Finally, we export ground-truth from origin image by setting the MCS pixel to 255 and the non MCS pixel to 0. As is common in remote sensing applications, MCSRSI is highly unbalanced, with the distribution of MCS regions in frame is less than 5\% (see Fig. \ref{fig5}).

\begin{figure*}
	\centering
        % input
		\subfigure{
             % \scriptsize{Input}
            % \rotatebox{90}{\scriptsize{~~~~~~~~~~~~~Input}}
		\includegraphics[width=0.15\textwidth]{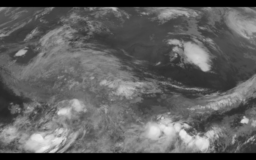}}
  		\subfigure{ 
		\includegraphics[width=0.15\textwidth]{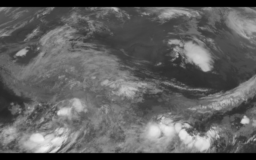}}
  		\subfigure{ 
		\includegraphics[width=0.15\textwidth]{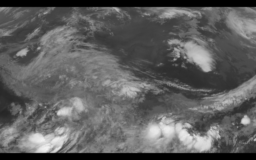}}
  		\subfigure{ 
		\includegraphics[width=0.15\textwidth]{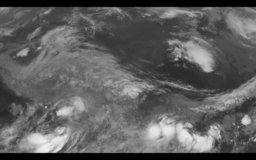}}
  		\subfigure{ 
		\includegraphics[width=0.15\textwidth]{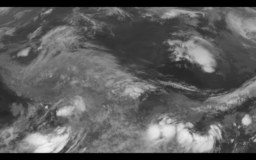}}
  		\subfigure{ 
		\includegraphics[width=0.15\textwidth]{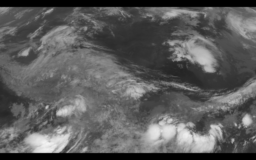}}

  % label
		\subfigure{
             % \scriptsize{Label}
            % \rotatebox{90}{\scriptsize{~~~~~~~~~~~~~Input}}
		\includegraphics[width=0.15\textwidth]{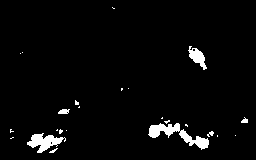}}
  		\subfigure{ 
		\includegraphics[width=0.15\textwidth]{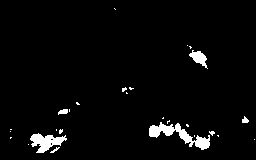}}
  		\subfigure{ 
		\includegraphics[width=0.15\textwidth]{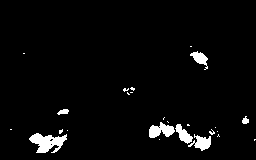}}
  		\subfigure{ 
		\includegraphics[width=0.15\textwidth]{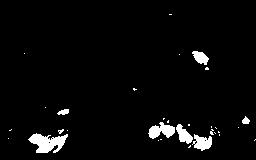}}
  		\subfigure{ 
		\includegraphics[width=0.15\textwidth]{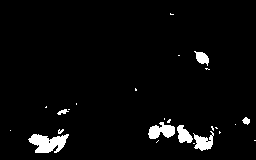}}
  		\subfigure{ 
		\includegraphics[width=0.15\textwidth]{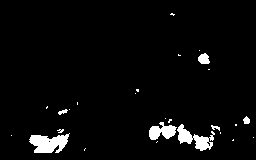}}

  		\subfigure[T=0]{
             % \scriptsize{Visual}
            % \rotatebox{90}{\scriptsize{~~~~~~~~~~~~~Input}}
		\includegraphics[width=0.15\textwidth]{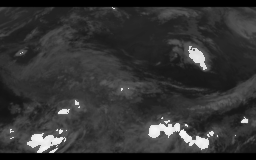}}
  		\subfigure[T=30]{ 
		\includegraphics[width=0.15\textwidth]{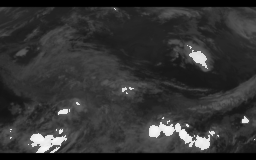}}
  		\subfigure[T=60]{ 
		\includegraphics[width=0.15\textwidth]{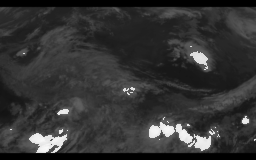}}
  		\subfigure[T=90]{ 
		\includegraphics[width=0.15\textwidth]{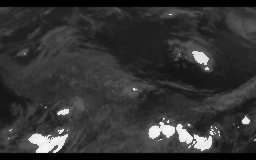}}
  		\subfigure[T=120]{ 
		\includegraphics[width=0.15\textwidth]{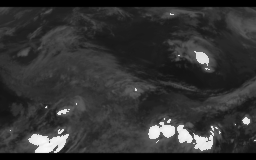}}
  		\subfigure[T=150]{ 
		\includegraphics[width=0.15\textwidth]{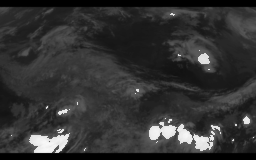}}
        \caption{A 6-frame-width time-series sequence in MCSRSI. From top to bottom, the order is as follows: input image, convective cloud label, and the distribution map of convective cloud.}
        \label{fig6}
\end{figure*}

To get the RSI sequence for MCS detection, we utilize a 6-frame-wide sliding window with 30 min interval to organize image into a time-series sequence (see Fig. \ref{fig6}). In total, we obtain 7857 time-series sequences. Due to the distinct seasonal characteristics of MCS regions distribution and to prevent overfitting, we sample each month's data separately. First, we partition the data for each month into 5 equal groups, with one group used as the test set and the remaining four groups used as the train set. Then, we utilize the sliding window approach to generate time-series sequence for each group separately. The sequences that are not continuous in time are discarded. Through this approach, we obtain a train set with 6350 sequences and a test set with 1507 sequences. Both the train and test set contain time-series sequence from the entire year.
~\\
~\\
\noindent \textbf{Extreme Condition Evaluation}
Mesoscale convective system, characterized by its abruptness and challenges in monitoring, poses significant threats to property and human lives. The unpredictability and rapid nature of convective events often leave authorities and individuals alike scrambling to respond, making it a particularly dangerous weather phenomenon. We hope that our model can achieve better performance in extreme conditions, especially in regions where MCS occurs frequently and in dense clusters. To this end, we have conducted an experiment to evaluate the performance of model across a diverse array of MCS distributions. The scope of this experiment is vast, encompassing a range of MCS occurrences spanning from the relatively rare 1\% to the more common 5\%.
~\\
~\\
\noindent \textbf{Experimental Setup}
We implement the proposed method and other models in comparison with the Pytorch framework and conduct experiments on a single RTX 3090 GPU. We train MCSDNet, using Adam with a learning rate of 0.001, a batch size of 8 time-series sequences with 6 images and ReduceLROnPlateau as a scheduler to adjust the learning rate.
~\\
~\\
\noindent \textbf{Evaluation Metrics}
We consider probability of detection (POD), False Alarm Ratio (FAR) and critical success index (CSI) as evalutaion metrics to compare MCSDNet with other methods. The evaluation metrics are formulated as
\begin{equation}
	\begin{aligned}
                POD=\frac{TP}{TP+FN}, 
	\end{aligned}
	\label{eq6}
\end{equation}
\begin{equation}
	\begin{aligned}
                FAR=\frac{FP}{TP+FP},
	\end{aligned}
	\label{eq7}
\end{equation}
\begin{equation}
	\begin{aligned}
                CSI=\frac{FP}{TP+FP+FN},
	\end{aligned} 
	\label{eq8}
\end{equation}
where TP is true positive, FP is false positive, and FN is false negative. POD is used to measure the proportion of correctly detected MCS regions in the MCS set. FAR is used to measure the proportion of misclassified samples among those classified as MCS. CSI represents the IoU of positive samples, indicating the proportion of overlap between the positive regions in the ground truth and the prediction. Since POD and FAR are always inversely proportional, it is difficult to measure the quality of the model using these two metrics alone. Therefore, CSI is the primary metric used for performance evaluation, with POD and FAR as secondary metrics. A higher CSI and POD, and a lower FAR, indicate better model performance.
\subsection{Performance Evaluation}
\begin{table*}
        \renewcommand\arraystretch{1.5}
	\caption{Experiment results on MCSRSI. The best results are highlighted in bold. O:the model targets on single-frame. M:the model targets on multi-frames. R:remote sensing image analysis methods. S:semantic segmentation methods. V:video understanding methods.}
	\begin{center}
		\smallskip\scalebox
		{1.0}{
			\begin{tabular}{c c c c c c}
				\hline
				% \multicolumn{1}{c|}{\multirow{1}{*}{Setting}}
                \multicolumn{1}{c|}{\multirow{1}{*}{Method}}
                & \multicolumn{1}{c|}{\multirow{1}{*}{Frames}}
                & \multicolumn{1}{c|}{\multirow{1}{*}{Category}}  
                & \multicolumn{1}{c|}{\multirow{1}{*}{POD $\uparrow$ }}
                &\multicolumn{1}{c|}{\multirow{1}{*}{FAR $\downarrow$}}
                & \multicolumn{1}{c}{\multirow{1}{*}{CSI $\uparrow$}} \\
                \hline \hline
                % Cloud Detection Method
                % \multirow{2}{*}{Cloud Detection Method}
                RS-NET \cite{jeppesen2019cloud} & O & R & 0.85966 & 0.12059 & 0.77399 \\

                CS-CNN \cite{dronner2018fast}& O & R & 0.85829  & 0.10719 & 0.78452 \\
                ConvectionUnet \cite{wang2023convection}& O & R & 0.86142  & 0.12234 & 0.77394 \\

                % Semantic Segmentation Methods Based on Single-Frame Images
                % \multirow{3}{*}{Semantic Segmentation Methods Based on Single-Frame Images}
                 Swin+UperNet \cite{liu2021swin}&O &S & 0.42113 & 0.25495 & 0.36962 \\

                DeepLabv3+ \cite{chen2018encoder}&O &S & 0.49900 & 0.21797 & 0.43248 \\

                Unet2D \cite{ronneberger2015u}&O &S & 0.86545 & 0.13145 & 0.77201 \\

                % \multirow{3}{*}{Semantic Segmentation Methods Based on Image Sequences}
                Video Swin \cite{liu2022video}&M &V & 0.44182 & 0.28698 & 0.37855 \\ 
                SimVP \cite{gao2022simvp} &M &V &0.77602 &0.16065 &0.73329 \\
                LSTMUnet &M &V& 0.90183  & 0.16943  &  0.76884 \\
                Unet3D &M &V & 0.87220 & 0.08692 & 0.81172 \\
                \hline
                 \textbf{MCSDNet} &\textbf{M} &\textbf{R}& \textbf{0.92537} & \textbf{0.06530} &\textbf{0.87162} \\
                \hline
		\end{tabular}}
	\end{center}
	\label{table2}
\end{table*}

We evaluate our proposed method against strong baselines, including RSI analysis methods (CS-CNN \cite{dronner2018fast}, RS-NET \cite{jeppesen2019cloud}), semantic segmentation methods (Unet2D \cite{ronneberger2015u}, DeepLabv3+ \cite{chen2018encoder}, Swin Transformer \cite{liu2021swin}), and video understanding (Unet3D, LSTMUnet, Video Swin \cite{liu2022video}, SimVP \cite{tan2022simvp}).

~\\
\noindent \textbf{Mesoscale Convective System Detection on MCSRSI Dataset}
Table \ref{table2} summarizes the results of baseline models and MCSDNet on MCSRSI dataset. The results show that our proposed MCSDNet achieves state-of-the-art performance on MCS detection task when comparing with other methods. To be fair, all models are without a pretrain process. We can see that RSI analysis methods outperform semantic segmentation methods. It is because RSI analysis models are specifically designed for remote sensing imagery. They are more effective at extracting features from MCS compared to general semantic segmentation models. In semantic segmentation models, Unet has better performance than DeepLabv3+ and Swin Transformer. Texture feature is significant for MCS detection. Unet uses skip connections to capture multi-scale spatial information, allowing it to extract texture information from the image better. On the contrary, Swin Transformer and DeepLabv3+  focus on capturing the semantic information of images. Therefore, although they perform well in natural image semantic segmentation, they are not suitable for remote sensing image. The video understanding methods have better performance than other methods. This demonstrates that introducing temporal information can effectively improve the performance of MCS detection.
Different from the above models, MCSDNet captures multi-scale spatiotemporal information from different encoder levels and apply a pyramid pooling to adaptively adjust the field-of-view. Then, our model utilizes DSTA to extract intra-frame features and inter-frame correlations. This not only emphasizes the texture feature of image but also learns the variation trend in the development process of MCS regions. MCSDNet integrates the advantages of RSI analysis models and video understanding models. By combining spatiotemporal data, it effectively improves the performance of MCS detection.

\begin{figure*}
	\centering
	% first line
	\subfigure[CS-CNN]{ 
		% \label{fig7a} %% label for first subfigure 
		\includegraphics[width=0.3\textwidth]{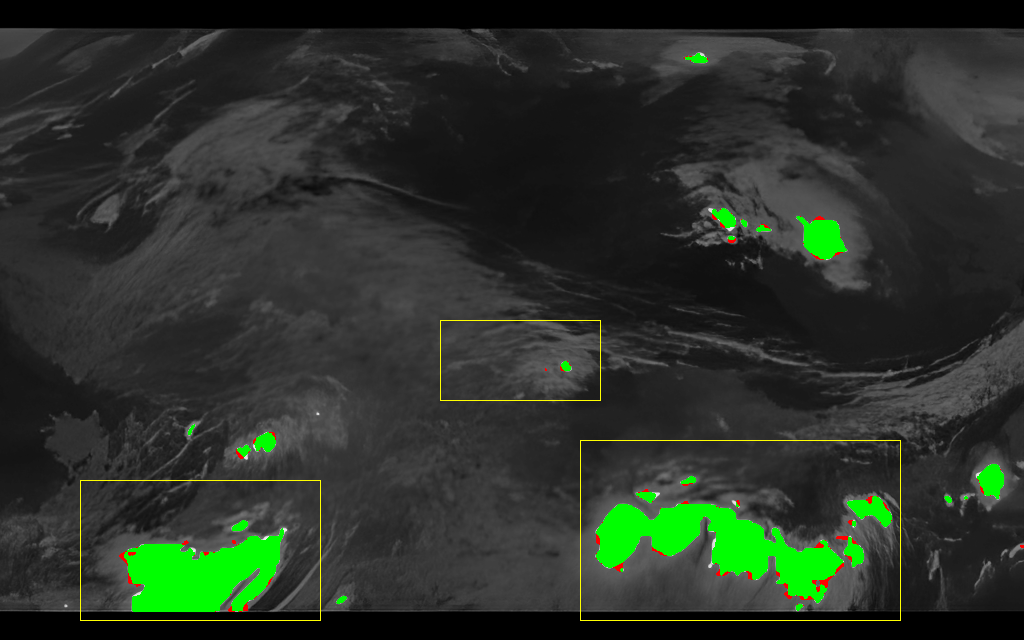}}
	\subfigure[RS-NET]{ 
		% \label{fig7b} %% label for first subfigure 
		\includegraphics[width=0.3\textwidth]{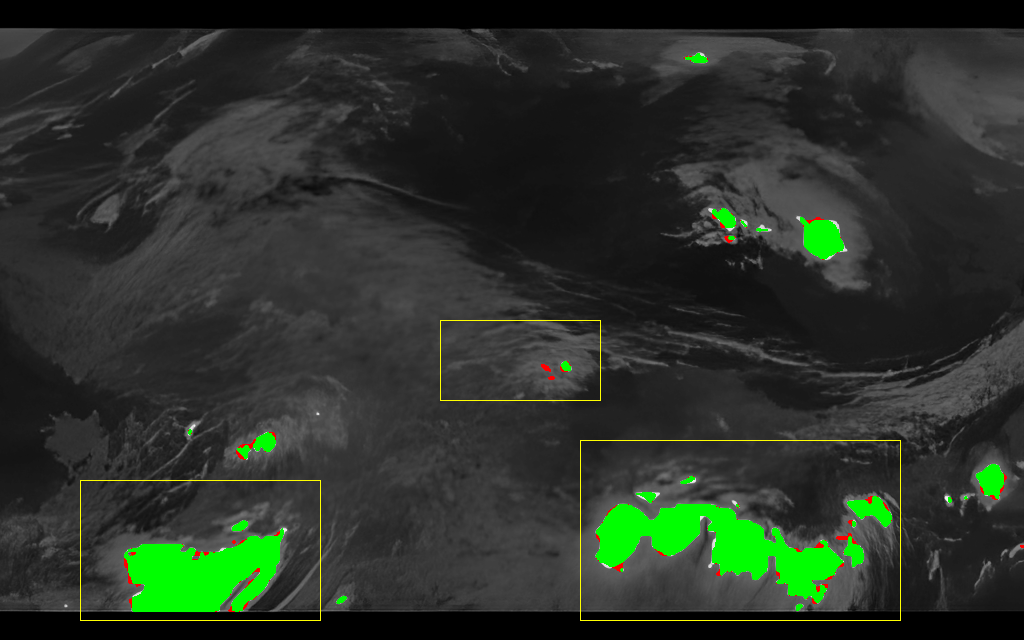}}
	\subfigure[Swin]{ 
		% \label{fig7b} %% label for first subfigure 
		\includegraphics[width=0.3\textwidth]{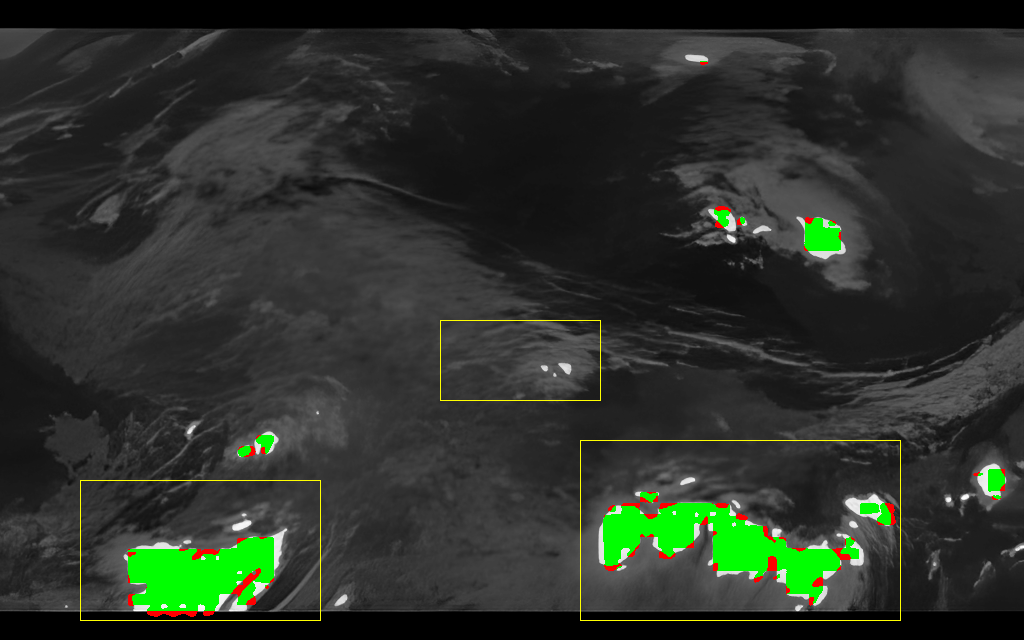}}

  % second line
  	\subfigure[DeepLabv3+]{ 
		% \label{fig7a} %% label for first subfigure 
		\includegraphics[width=0.3\textwidth]{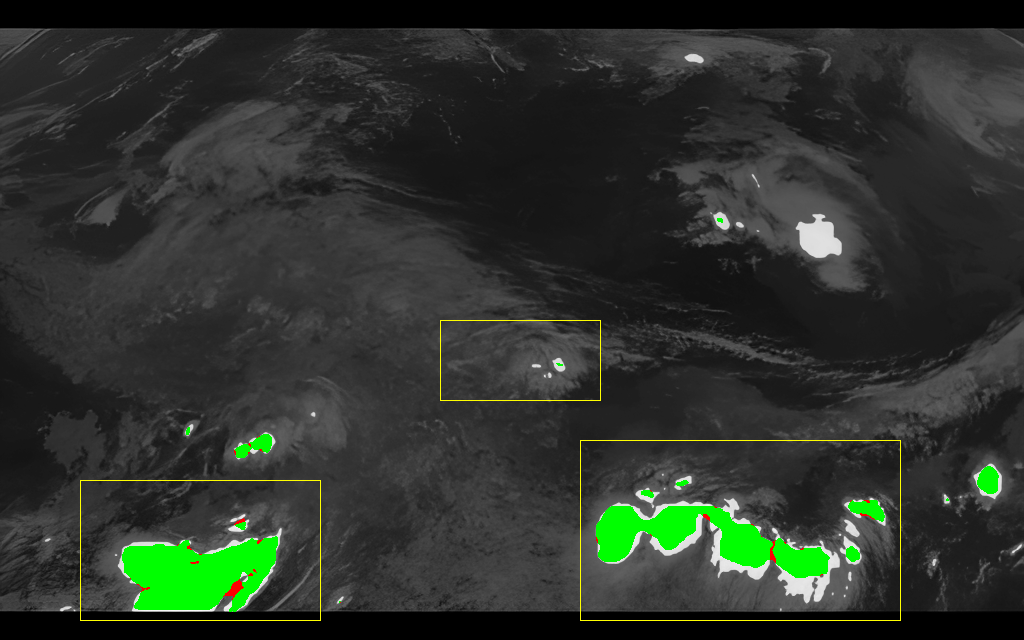}}
	\subfigure[Unet2D]{ 
		% \label{fig7b} %% label for first subfigure 
		\includegraphics[width=0.3\textwidth]{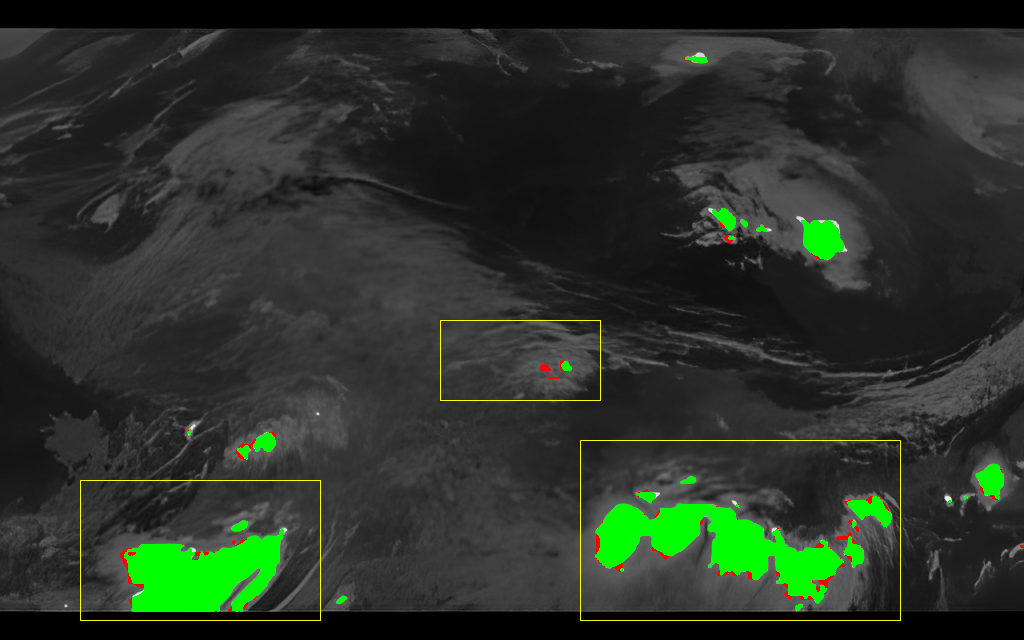}}
	\subfigure[Video Swin]{ 
		% \label{fig7b} %% label for first subfigure 
		\includegraphics[width=0.3\textwidth]{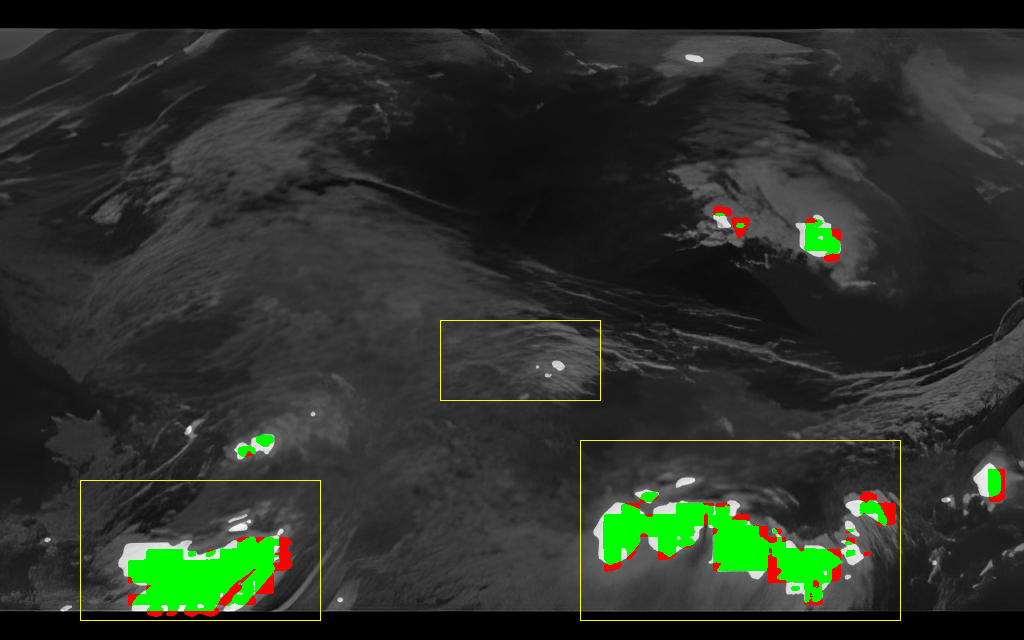}}

    % third line
  	\subfigure[LSTMUnet]{ 
		% \label{fig7a} %% label for first subfigure 
		\includegraphics[width=0.3\textwidth]{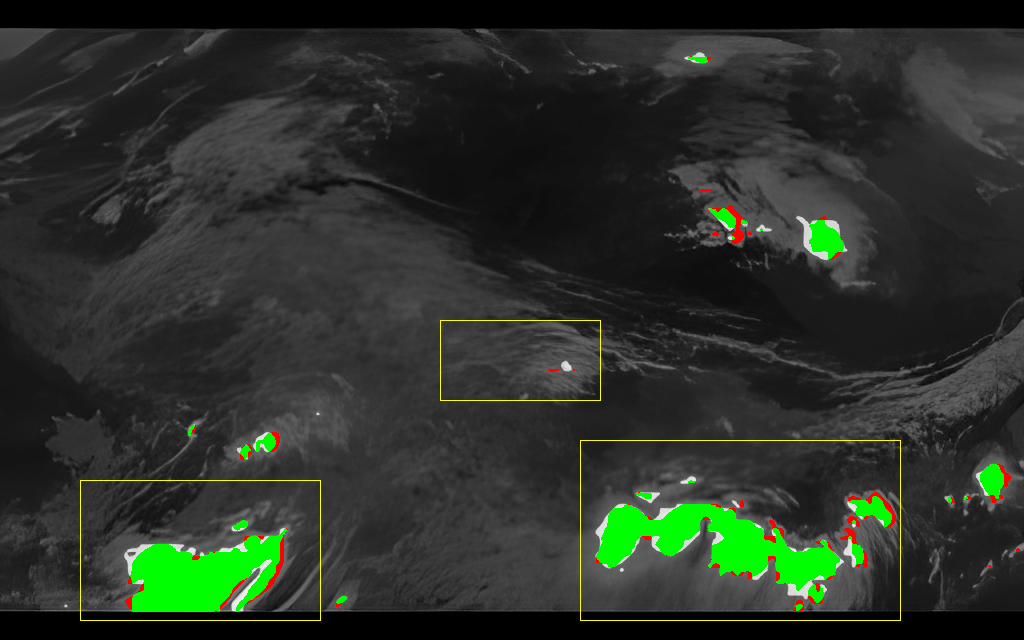}}
	\subfigure[Unet3D]{ 
		% \label{fig7b} %% label for first subfigure 
		\includegraphics[width=0.3\textwidth]{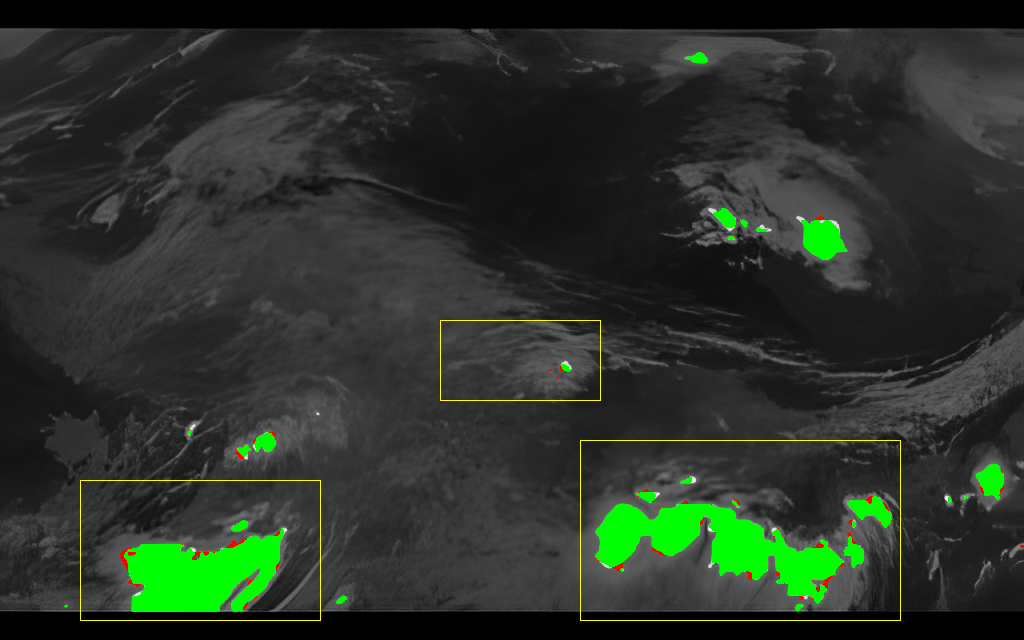}}
	\subfigure[MCSDNet]{ 
		% \label{fig7b} %% label for first subfigure 
		\includegraphics[width=0.3\textwidth]{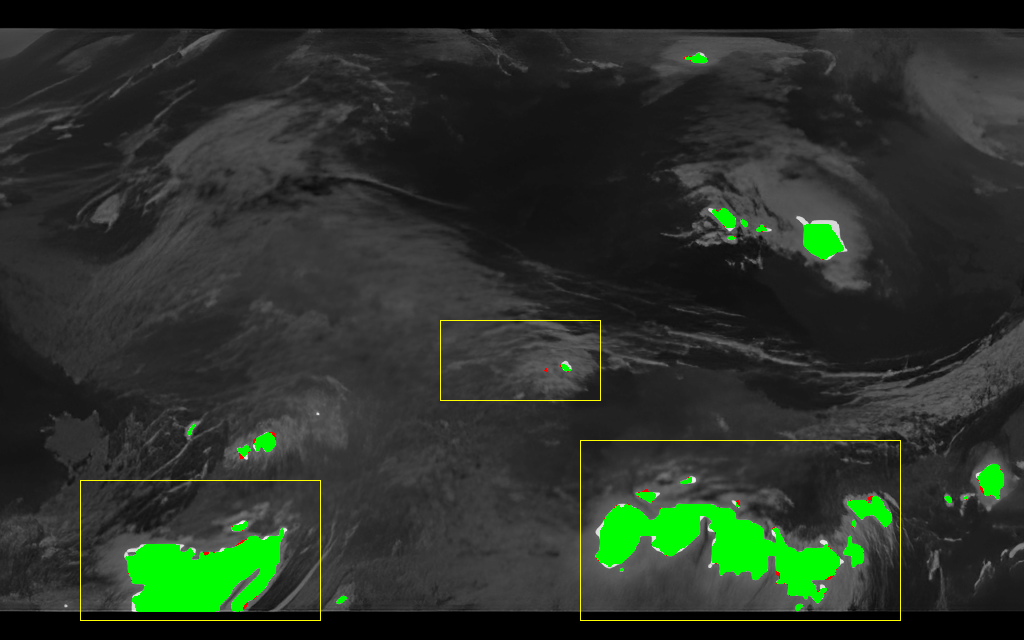}}
          \caption{Visual comparisons of different methods in MCSRSI. The white area represents convective cloud and the black area is non convective cloud. Besides, the green area represents true positive detection, while the red area represents false negative detection. (a) Result of CS-CNN (b) Result of RS-NET (c) Result of Swin Transformer (d) Result of DeepLabv3+ (e) Result of Unet2D (f) Result of Video Swin (g) Result of LSTMUnet (h) Result of Unet3D (i) Result of MCSDNet}
        \label{fig7}
\end{figure*}

Fig. \ref{fig7} shows the visual comparison of MCSDNet and other methods on the example from MCSRSI. Black represents the background region, white represents the MCS region, green represents the correctly detected region, and red represents the region classified as convective cloud when it is not. From the results, we can see that MCSDNet has better performance than other methods. It has more true positive and less false positive detection. The intra-frame features and inter-frame correlations enable the model to capture long-term relations and improve the performance of MCS detection.
\begin{figure*}
	\centering
	\subfigure[Probability of Detection(POD)]{ 
		\label{fig9a} %% label for first subfigure 
		\includegraphics[width=.32\textwidth]{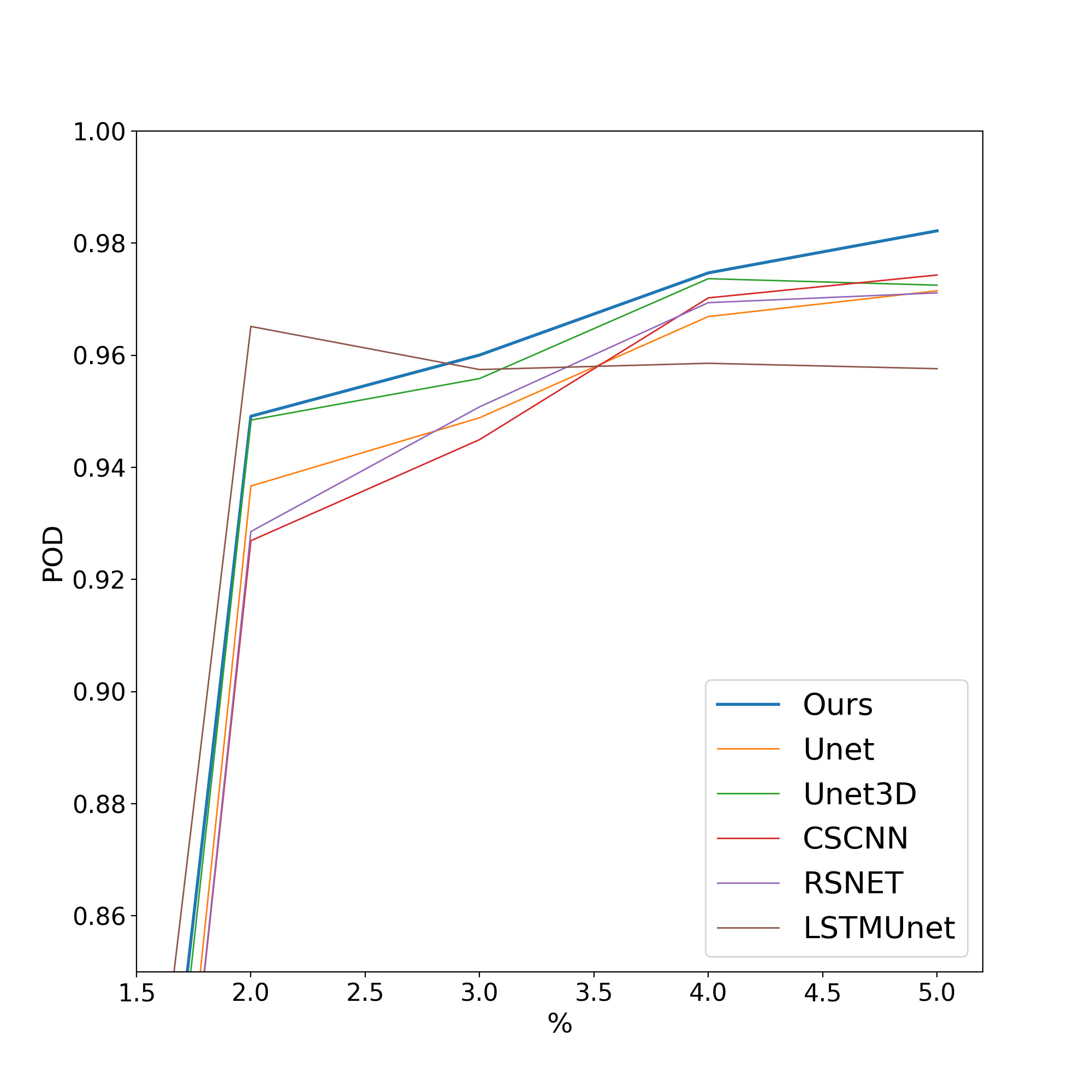}}
	\subfigure[False Alarm Ratio(FAR)]{ 
		\label{fig9b} %% label for first subfigure 
		\includegraphics[width=.32\textwidth]{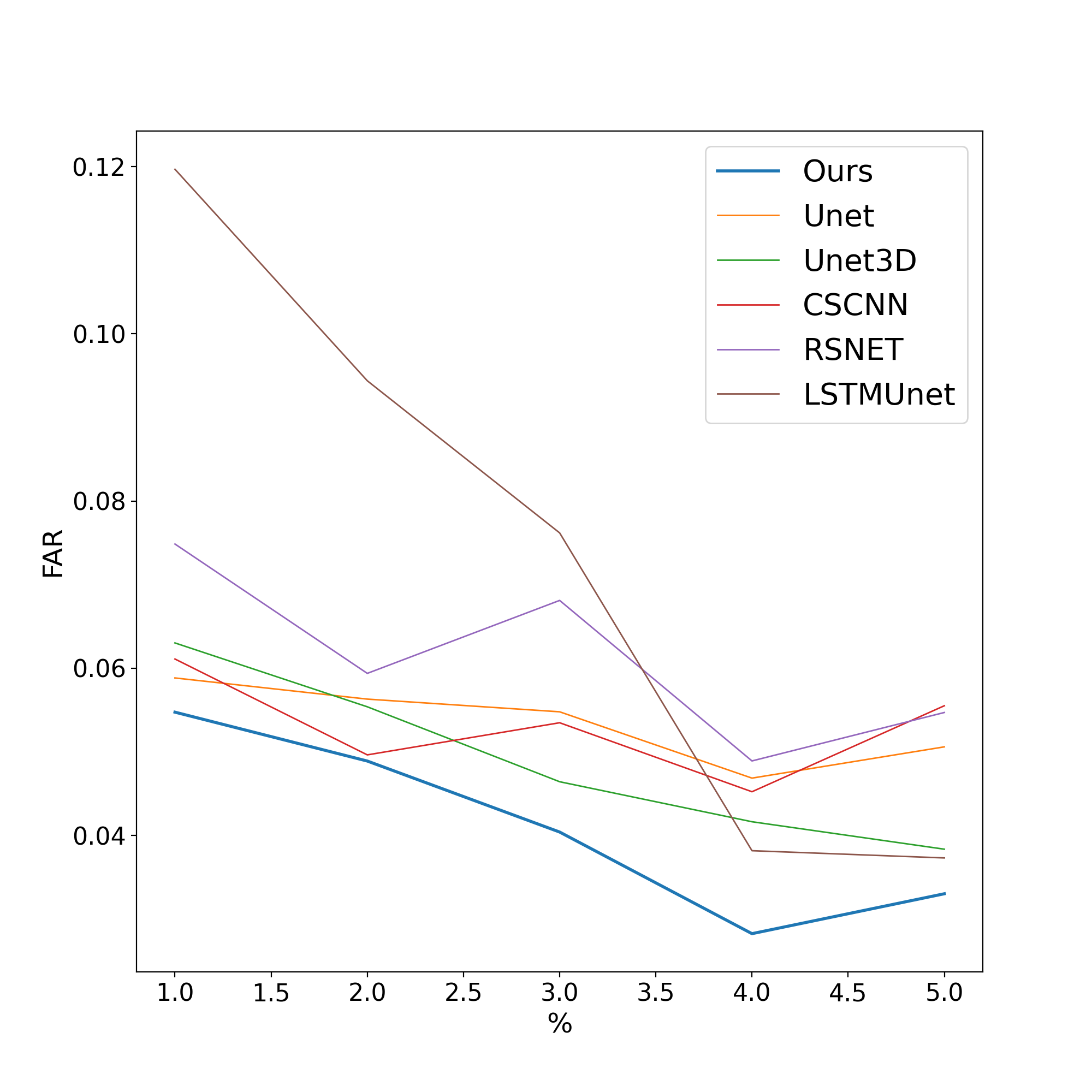}}
	\subfigure[Critical Success Index(CSI)]{ 
		\label{fig9c} %% label for first subfigure 
		\includegraphics[width=.32\textwidth]{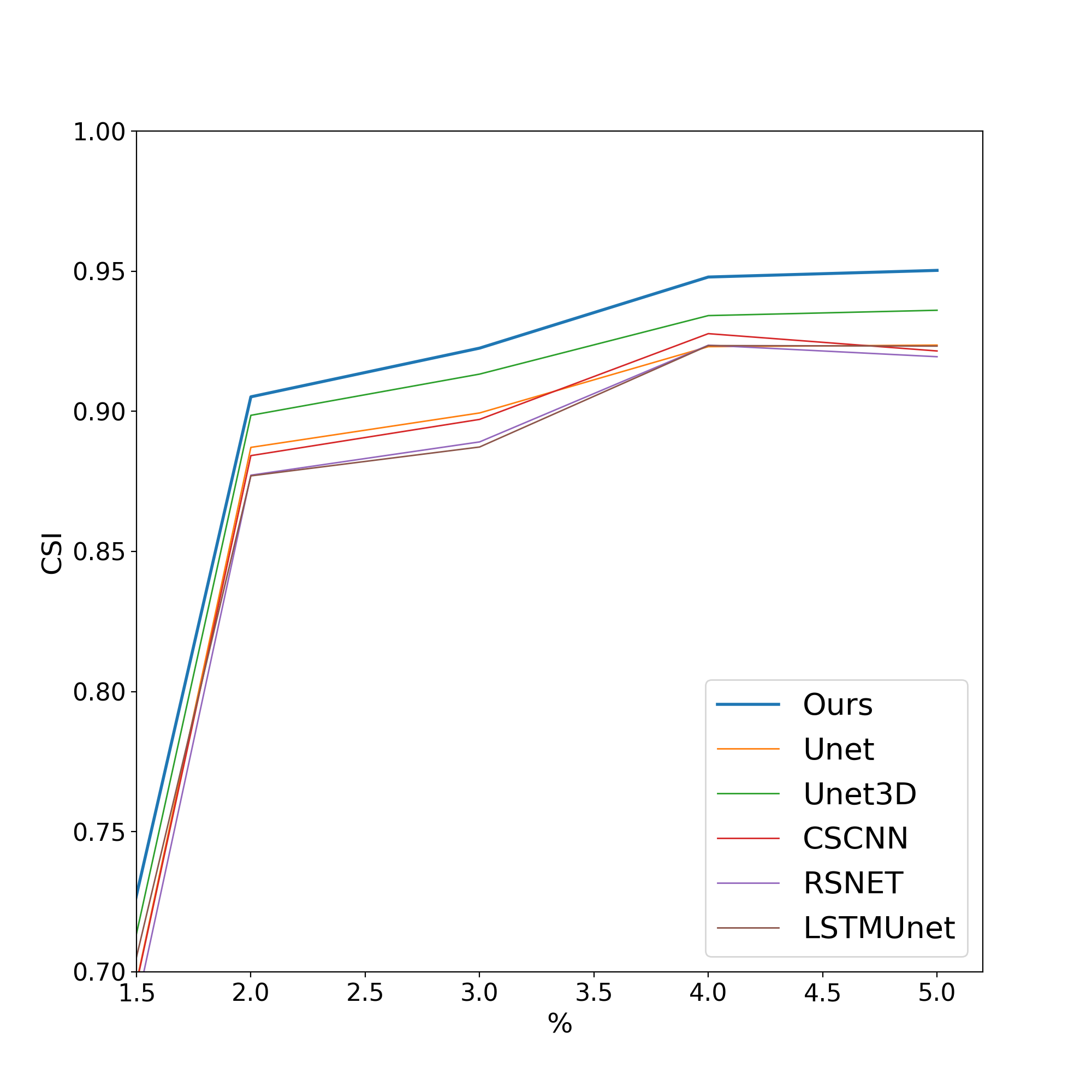}}
	\caption{Performance curves of the models under different convective weather distributions. The blue line represents our proposed model.}
	\label{fig9}
\end{figure*}
% \begin{figure}
% 	\centering
% 	\includegraphics[width=\columnwidth]{image/extreme/pod.png}
% 	\caption{Probability of Detection(POD)}
% 	\label{fig9}
% \end{figure}
% \begin{figure}
% 	\centering
%     \includegraphics[width=\columnwidth]{image/extreme/far.png}
% 	\caption{False Alarm Ratio(FAR)}
% 	\label{fig10}
% \end{figure}
% \begin{figure}
% 	\centering
%     \includegraphics[width=\columnwidth]{image/extreme/csi.png}
% 	\caption{Critical Success Index(CSI)}
% 	\label{fig11}
% \end{figure}
~\\

\noindent \textbf{Dense Distribution of Mesoscale Convective System} Fig. \ref{fig9} presents a comprehensive comparison of the performance of our proposed model and the baseline model across various MCS distributions. All models are trained on MCSRSI with the same experiment setup. The results clearly illustrate the superiority of our model, especially in conditions characterized by frequent and dense convective weather events. 

As the distribution of MCS regions increases, our model demonstrates a consistent and significant improvement in performance. Compared to the baseline models, our model gains better performance in extreme conditions, particularly when MCS regions occur frequently and densely. For example, when the distribution of MCS regions is 5\%, our model achieves the best performance in all metrics. With a CSI of 0.95, our model demonstrates a high level of accuracy in identifying convective weather events. Additionally, a low FAR of 0.03 minimizes the occurrence of unnecessary warnings, while a high POD of 0.98 ensures that the majority of convective weather events are captured and not missed. This improvement can be attributed to the introduction of spatial and temporal information, which plays a crucial role in enhancing the model's ability to accurately detect MCS regions patterns and makes our model gain better performance in extreme conditions.

% \begin{figure}
% 	\centering
	
% 	\subfigure[Performance]{ 
% 		\label{fig5a} %% label for first subfigure 
% 		\includegraphics[width=0.47\columnwidth]{visual_a.eps}}
% 	\subfigure[Feature space]{ 
% 		\label{fig5b} %% label for first subfigure 
% 		\includegraphics[width=0.47\columnwidth]{visual_b.eps}}
% 	\caption{Visualization of Neural ODE on miniImagenet.}
% 	\label{fig99}
% \end{figure}

\subsection{Ablation Studies}
% dual spatiotemporal attention ablation
\begin{table}
        \renewcommand\arraystretch{1.5}
	\caption{The role of multi-scale spatiotemporal information}
	\begin{center}
		\smallskip\scalebox
		{1.0}{
		\begin{tabular}{c c c c}
				\hline
                \multicolumn{1}{c}{\multirow{1}{*}{Multi-scale}}
                & \multicolumn{1}{c}{\multirow{1}{*}{POD $\uparrow$ }}
                &\multicolumn{1}{c}{\multirow{1}{*}{FAR $\downarrow$}}
                & \multicolumn{1}{c}{\multirow{1}{*}{CSI $\uparrow$}} \\
                \hline \hline
                % Cloud Detection Method
                  & 0.88633  & 0.06031 & 0.84078 \\
                \hline
                 \checkmark & \textbf{0.92537}  & \textbf{0.06530} & \textbf{0.87162} \\
                \hline
		\end{tabular}}
	\end{center}
	\label{table55}
\end{table}

\label{section_4_5}
We conduct ablation studies to evaluate the performance of MCSDNet and the proposed modules inside it. We train the models on the train set of MCSRSI and then test them on the test set. For performance evaluation, we compare the performance metrics such as POD, FAR and CSI.
\begin{table}
        \renewcommand\arraystretch{1.5}
	\caption{The role of T-MSA and S-MSA in DSTA}
	\begin{center}
		\smallskip\scalebox
		{1.0}{
		\begin{tabular}{c c c c c}
				\hline
                \multicolumn{1}{c}{\multirow{1}{*}{Temporal Attention}}
                &\multicolumn{1}{c}{\multirow{1}{*}{Spatial Attention}}
                & \multicolumn{1}{c}{\multirow{1}{*}{POD $\uparrow$ }}
                &\multicolumn{1}{c}{\multirow{1}{*}{FAR $\downarrow$}}
                & \multicolumn{1}{c}{\multirow{1}{*}{CSI $\uparrow$}} \\
                \hline \hline
                % Cloud Detection Method
                 & &0.81650 &0.32835 &0.59543 \\
                 \hline
                 \checkmark &  & 0.84339 & 0.10359 & 0.78091 \\
                \hline
                 & \checkmark & 0.89667 & 0.09054 & 0.83123 \\
                \hline
                \checkmark & \checkmark & \textbf{0.92537} & \textbf{0.06530} & \textbf{0.87162} \\
                \hline
		\end{tabular}}
	\end{center}
	\label{table4}
\end{table}
~\\
~\\
\noindent \textbf{What's the role of multi-scale spatiotemporal information?}
Due to MCS detection is a dense prediction task, we desire more detail spatiotemporal information to segment objects accurately. In MCSDNet, we concatenate the feature maps from different encoder levels and then apply a pyramid pooling to capture multi-scale spatiotemporal information. As shown in Table. \ref{table55}, the model with multi-scale spatiotemporal information gains better performance than that with single-scale information, with a 4\% improvement in POD, a 3\% decrease in FAR and a 3\% enhancement in CSI. It shows that introducing multi-scale information can significantly improve the accuracy of MCS detection. 
\begin{table}
        \renewcommand\arraystretch{1.5}
	\caption{Performance of different STMU.}
	\begin{center}
		\smallskip\scalebox
		{1.0}{
		\begin{tabular}{c c c c}
				\hline
                \multicolumn{1}{c}{\multirow{1}{*}{STMU}}
                & \multicolumn{1}{c}{\multirow{1}{*}{POD $\uparrow$ }}
                &\multicolumn{1}{c}{\multirow{1}{*}{FAR $\downarrow$}}
                & \multicolumn{1}{c}{\multirow{1}{*}{CSI $\uparrow$}} \\
                \hline \hline
                % Cloud Detection Method
                 None & 0.81650 & 0.32835 & 0.59543 \\
                 Conv3D & 0.78002 & 0.28656 & 0.61255 \\
                ConvLSTM & 0.95834  & 0.09352 & 0.87333 \\
                Vision Transformer& 0.89667  & 0.09054 & 0.83123 \\
                \hline
                DSTA & 0.92537  & 0.06530 & 0.87162 \\
                \hline
		\end{tabular}}
	\end{center}
	\label{table66}
\end{table}
~\\
~\\
\noindent \textbf{Can we use other spatiotemporal modules as STMU?}
In MCSDNet, we utilize STMU to capture the long-range temporal evolution of MCS regions from RSI sequence. With the rapid development of spatiotemporal modules, researchers maybe confused to choose which one as STMU \cite{gao2022simvp}. In Table. \ref{table66}, we explore the performance of MCSDNet with different STMU, \emph{e.g.}, CNN, RNN, Transformer and DSTA. From the results, we can see that the models with STMU detect MCS regions more accurately. Our proposed model exhibits remarkable scalability, facilitating the seamless integration of various spatiotemporal modules as STMU. Besides, MCSDNet maintains the original performance standards of the spatiotemporal modules without any compromise. 

We design a new spatiotemporal modules named DSTA as STMU. Different from the previous modules, our proposed DSTA captures both intra-frame features and inter-frame correlations, and achieves the best performance among all STMUs. 
~\\
~\\
\noindent \textbf{What's the role of attention modules in DSTA?}
In our implementation, MCSDNet employs DSTA as STMU to capture spatiotemporal information of MCS regions. DSTA consists of a temporal attention module(T-MSA) and a spatial attention module(S-MSA). As shown in Table \ref{table4}, when comparing with the baseline model, the models with attention modules improve the performance remarkably. The model with only T-MSA outperforms the baseline by 18.55\% in CSI and  22.48\% in FAR. T-MSA aims to capture the temporal evolution of RSI sequence and locate the MCS regions with cross-frames temporal consistency, which improves the FAR of MCS detection task. Meanwhile, employing S-MSA individually improves the detection accuracy from 81.65\% to 89.67\% in POD and 59.54\% to 86.86\% in CSI. S-MSA can emphasizes the texture features and distinguish MCS regions from none MCS regions better. When we integrate the two attention modules together, the model with both T-MSA and S-MSA achieves the best performance in all evaluation metrics, with 92.54\% in POD, 6.53\% in FAR and 87.16\% in CSI. From the results, we can see that introducing spatiotemporal information is significant to improve the performance of MCS detection task.
% time interval Ablation
\begin{table}
        \renewcommand\arraystretch{1.5}
	\caption{The optimal time interval in RSI sequence.}
	\begin{center}
		\smallskip\scalebox
		{1.0}{
		\begin{tabular}{c c c c}
				\hline
                \multicolumn{1}{c}{\multirow{1}{*}{Interval}}
                & \multicolumn{1}{c}{\multirow{1}{*}{POD $\uparrow$ }}
                &\multicolumn{1}{c}{\multirow{1}{*}{FAR $\downarrow$}}
                & \multicolumn{1}{c}{\multirow{1}{*}{CSI $\uparrow$}} \\
                \hline \hline

                % Cloud Detection Method
                 15 min & 0.85548 & 0.09623 & 0.79255 \\
                \hline
                 30min & \textbf{0.92537}  & \textbf{0.06530} & \textbf{0.87162} \\
                \hline
                
                 60min & 0.86778 & 0.09234 & 0.80540 \\
                 \hline
                 120min & 0.85667 & 0.09510 & 0.79666 \\
                \hline
		\end{tabular}}
	\end{center}
	\label{table5}
\end{table}
~\\
~\\
\noindent \textbf{What is the optimal time interval for image sequence?} It is important for MCS detection to introduce spatiotemporal information. We should choose appropriate time interval when designing RSI sequence. If time interval is too small, the changes in MCS regions are not significant and we can not effectively capture temporal evolution. Conversely, when the time interval is too large, there is a lack of correlations between consecutive frames. We conduct ablation study in four time interval: 15min, 30min, 60min and 120min. As shown in Table \ref{table5}, when the time interval is 30min, MCSDNet achieve the best performance with 92.54\% in POD, 6.53\% in FAR and 87.16\% in CSI. From the results, we find that time interval in RSI sequence significantly affects the performance of MCSDNet. As the time interval increases, the performance of model first improves and then declines. This suggests that both too small and too large time interval are not conducive to capture intra-frame and inter-frame variation. 
\begin{table}
        \renewcommand\arraystretch{1.5}
	\caption{Performance of network with combination of the individual module.}
	\begin{center}
		\smallskip\scalebox
		{1.0}{
		\begin{tabular}{c c c c c c}
				\hline
                \multicolumn{1}{c}{\multirow{1}{*}{Backbone}}
                & \multicolumn{1}{c}{\multirow{1}{*}{STMU}}
                & \multicolumn{1}{c}{\multirow{1}{*}{Decoder}}
                & \multicolumn{1}{c}{\multirow{1}{*}{POD $\uparrow$ }}
                &\multicolumn{1}{c}{\multirow{1}{*}{FAR $\downarrow$}}
                & \multicolumn{1}{c}{\multirow{1}{*}{CSI $\uparrow$}} \\
                \hline \hline
                % Cloud Detection Method
                 \checkmark & \checkmark &  & 0.08112  & 0.37670 & 0.07539\\
                \hline
                 \checkmark &  & \checkmark & 0.81650 & 0.32835 & 0.59543 \\
                \hline
                 \checkmark & \checkmark & \checkmark & \textbf{0.92537} & \textbf{0.06530} & \textbf{0.87162} \\
                \hline
		\end{tabular}}
	\end{center}
	\label{table8}
\end{table} 
~\\
\noindent \textbf{What roles do the Encoder, STMU and Decoder play?} The proposed MCSDNet consists of encoder, STMU, and decoder. Although our model has a simple architecture, it can improve the performance of MCS detection remarkably. We are eager to know the modules in MCSDNet how to influence the performance of MCS detection. As shown in Table \ref{table8}, we find that STMU effectively improves the performance of model for MCS detection. Without STMU, the performance of model drops by 10.89\% in POD, 26.31\% in FAR and 27.62\%. It is important for MCS detection to introduce spatiotemporal information. In MCSDNet, the encoder downsamples the origin image and extracts the spatial and texture information of MCS, and then the decoder gradually restores the feature map from encoder to the size of input frame. Besides, with skip connection, the lost spatial information during downsampling is compensated to the input of each decoder layer. We see that the performance of model without decoder significantly deteriorates so that it can not complete the MCS detection task. STMU and Decoder are essential for model to perform well on MCS detection.

\section{Conclusion}
\label{section_5}
In this paper, we have proposed a novel deep learning model MCSDNet for MCS detection. MCSDNet has a simple architecture and is easy to expand. Different from previous methods, MCSDNet utilizes the multi-scale spatiotemporal information for MCS detection for the first time, which is more suitable to extreme conditions, especially when convective weather occurs frequently. In MCSDNet, we use STMU to capture temporal  evolution in life cycle of MCS. STMU is a scalable module, which can be replaced by other spatiotemporal modules. It means that the future works about spatiotemporal modules can be easily migrated to our model. Besides, we design a new spatiotemporal modules named DSTA, which can capture both intra-frame features and inter-frame correlations. Then, we create a new dataset MCSRSI, the first large-scale dataset for MCS detection based on visible channel image. Evaluated on MCSRSI, our proposed MCSDNet gains state-of-the-art performance over existing MCS detection, semantic segmentation and video understanding methods. We hope that the combination of our open-access dataset and promising results will encourage the future research for MCS detection task, whose economic and environmental stakes can not be understated.   

% \section*{Acknowledgments}
% This work was supported by the Shenzhen Science and Technology Program under Grant No. JCYJ201805071-83823045, JCYJ20200109113014456, and JCYJ20210324-120208022, and NSFC under Grant No. 61972111. 

\ifCLASSOPTIONcaptionsoff
  \newpage
\fi



%\bibliographystyle{IEEEtran}
%\bibliography{egbib}

\bibliographystyle{IEEEtran}
\bibliography{main}

\begin{thebibliography}{10}
\providecommand{\url}[1]{#1}
\csname url@samestyle\endcsname
\providecommand{\newblock}{\relax}
\providecommand{\bibinfo}[2]{#2}
\providecommand{\BIBentrySTDinterwordspacing}{\spaceskip=0pt\relax}
\providecommand{\BIBentryALTinterwordstretchfactor}{4}
\providecommand{\BIBentryALTinterwordspacing}{\spaceskip=\fontdimen2\font plus
\BIBentryALTinterwordstretchfactor\fontdimen3\font minus \fontdimen4\font\relax}
\providecommand{\BIBforeignlanguage}[2]{{%
\expandafter\ifx\csname l@#1\endcsname\relax
\typeout{** WARNING: IEEEtran.bst: No hyphenation pattern has been}%
\typeout{** loaded for the language `#1'. Using the pattern for}%
\typeout{** the default language instead.}%
\else
\language=\csname l@#1\endcsname
\fi
#2}}
\providecommand{\BIBdecl}{\relax}
\BIBdecl

\bibitem{fiolleau2013algorithm}
T.~Fiolleau and R.~Roca, ``An algorithm for the detection and tracking of tropical mesoscale convective systems using infrared images from geostationary satellite,'' \emph{IEEE transactions on Geoscience and Remote Sensing}, vol.~51, no.~7, pp. 4302--4315, 2013.

\bibitem{wang2023convection}
Y.~Wang and B.~Xiao, ``Convection-unet: A deep convolutional neural network for convection detection based on the geo high-speed imager of fengyun-4b,'' in \emph{2023 International Conference on Pattern Recognition, Machine Vision and Intelligent Algorithms (PRMVIA)}.\hskip 1em plus 0.5em minus 0.4em\relax IEEE, 2023, pp. 163--168.

\bibitem{yang2023convective}
Y.~Yang, C.~Zhao, Y.~Sun, Y.~Chi, and H.~Fan, ``Convective cloud detection and tracking using the new-generation geostationary satellite over south china,'' \emph{IEEE Transactions on Geoscience and Remote Sensing}, 2023.

\bibitem{zhu2012object}
Z.~Zhu and C.~E. Woodcock, ``Object-based cloud and cloud shadow detection in landsat imagery,'' \emph{Remote sensing of environment}, vol. 118, pp. 83--94, 2012.

\bibitem{zuo2022identification}
Y.~Zuo, Z.~Hu, S.~Yuan, J.~Zheng, X.~Yin, and B.~Li, ``Identification of convective and stratiform clouds based on the improved dbscan clustering algorithm,'' \emph{Advances in Atmospheric Sciences}, vol.~39, no.~12, pp. 2203--2212, 2022.

\bibitem{utsav2017statistical}
B.~Utsav, S.~M. Deshpande, S.~K. Das, and G.~Pandithurai, ``Statistical characteristics of convective clouds over the western ghats derived from weather radar observations,'' \emph{Journal of Geophysical Research: Atmospheres}, vol. 122, no.~18, pp. 10--050, 2017.

\bibitem{le2017deep}
M.~Le~Goff, J.-Y. Tourneret, H.~Wendt, M.~Ortner, and M.~Spigai, ``Deep learning for cloud detection,'' in \emph{8th International Conference of Pattern Recognition Systems (ICPRS 2017)}.\hskip 1em plus 0.5em minus 0.4em\relax IET, 2017, pp. 1--6.

\bibitem{li2019deep}
S.~Li, W.~Song, L.~Fang, Y.~Chen, P.~Ghamisi, and J.~A. Benediktsson, ``Deep learning for hyperspectral image classification: An overview,'' \emph{IEEE Transactions on Geoscience and Remote Sensing}, vol.~57, no.~9, pp. 6690--6709, 2019.

\bibitem{10153685}
S.~K. Roy, A.~Deria, D.~Hong, B.~Rasti, A.~Plaza, and J.~Chanussot, ``Multimodal fusion transformer for remote sensing image classification,'' \emph{IEEE Transactions on Geoscience and Remote Sensing}, vol.~61, pp. 1--20, 2023.

\bibitem{10497695}
Y.~Huang, J.~Peng, G.~Zhang, W.~Sun, N.~Chen, and Q.~Du, ``Adversarial domain adaptation network with calibrated prototype and dynamic instance convolution for hyperspectral image classification,'' \emph{IEEE Transactions on Geoscience and Remote Sensing}, pp. 1--1, 2024.

\bibitem{basaeed2016supervised}
E.~Basaeed, H.~Bhaskar, and M.~Al-Mualla, ``Supervised remote sensing image segmentation using boosted convolutional neural networks,'' \emph{Knowledge-Based Systems}, vol.~99, pp. 19--27, 2016.

\bibitem{mohajerani2019cloud}
S.~Mohajerani and P.~Saeedi, ``Cloud-net: An end-to-end cloud detection algorithm for landsat 8 imagery,'' in \emph{IGARSS 2019-2019 IEEE International Geoscience and Remote Sensing Symposium}.\hskip 1em plus 0.5em minus 0.4em\relax IEEE, 2019, pp. 1029--1032.

\bibitem{yang2019cdnet}
J.~Yang, J.~Guo, H.~Yue, Z.~Liu, H.~Hu, and K.~Li, ``Cdnet: Cnn-based cloud detection for remote sensing imagery,'' \emph{IEEE Transactions on Geoscience and Remote Sensing}, vol.~57, no.~8, pp. 6195--6211, 2019.

\bibitem{10309842}
C.~Luo, Z.~Zhang, H.~Lin, B.~Zhang, X.~Li, T.~Zhang, and Y.~Ye, ``A practical online incremental learning framework for precipitation nowcasting,'' \emph{IEEE Transactions on Geoscience and Remote Sensing}, vol.~62, pp. 1--14, 2024.

\bibitem{10403855}
Z.~Zhao, X.~Dong, Y.~Wang, and C.~Hu, ``Advancing realistic precipitation nowcasting with a spatiotemporal transformer-based denoising diffusion model,'' \emph{IEEE Transactions on Geoscience and Remote Sensing}, vol.~62, pp. 1--15, 2024.

\bibitem{9686686}
X.~He, Y.~Zhou, J.~Zhao, D.~Zhang, R.~Yao, and Y.~Xue, ``Swin transformer embedding unet for remote sensing image semantic segmentation,'' \emph{IEEE Transactions on Geoscience and Remote Sensing}, vol.~60, pp. 1--15, 2022.

\bibitem{10497698}
L.~Lv and L.~Zhang, ``Advancing data-efficient exploitation for semi-supervised remote sensing images semantic segmentation,'' \emph{IEEE Transactions on Geoscience and Remote Sensing}, pp. 1--1, 2024.

\bibitem{shi2015convolutional}
X.~Shi, Z.~Chen, H.~Wang, D.-Y. Yeung, W.-K. Wong, and W.-c. Woo, ``Convolutional lstm network: A machine learning approach for precipitation nowcasting,'' \emph{Advances in neural information processing systems}, vol.~28, 2015.

\bibitem{dosovitskiy2020image}
A.~Dosovitskiy, L.~Beyer, A.~Kolesnikov, D.~Weissenborn, X.~Zhai, T.~Unterthiner, M.~Dehghani, M.~Minderer, G.~Heigold, S.~Gelly \emph{et~al.}, ``An image is worth 16x16 words: Transformers for image recognition at scale,'' \emph{arXiv preprint arXiv:2010.11929}, 2020.

\bibitem{tan2023temporal}
C.~Tan, Z.~Gao, L.~Wu, Y.~Xu, J.~Xia, S.~Li, and S.~Z. Li, ``Temporal attention unit: Towards efficient spatiotemporal predictive learning,'' in \emph{Proceedings of the IEEE/CVF Conference on Computer Vision and Pattern Recognition}, 2023, pp. 18\,770--18\,782.

\bibitem{schumacher2005organization}
R.~S. Schumacher and R.~H. Johnson, ``Organization and environmental properties of extreme-rain-producing mesoscale convective systems,'' \emph{Monthly weather review}, vol. 133, no.~4, pp. 961--976, 2005.

\bibitem{haberlie2019radar}
A.~M. Haberlie and W.~S. Ashley, ``A radar-based climatology of mesoscale convective systems in the united states,'' \emph{Journal of Climate}, vol.~32, no.~5, pp. 1591--1606, 2019.

\bibitem{chen2019mesoscale}
D.~Chen, J.~Guo, D.~Yao, Y.~Lin, C.~Zhao, M.~Min, H.~Xu, L.~Liu, X.~Huang, T.~Chen \emph{et~al.}, ``Mesoscale convective systems in the asian monsoon region from advanced himawari imager: Algorithms and preliminary results,'' \emph{Journal of Geophysical Research: Atmospheres}, vol. 124, no.~4, pp. 2210--2234, 2019.

\bibitem{huang2018long}
X.~Huang, C.~Hu, X.~Huang, Y.~Chu, Y.-h. Tseng, G.~J. Zhang, and Y.~Lin, ``A long-term tropical mesoscale convective systems dataset based on a novel objective automatic tracking algorithm,'' \emph{Climate dynamics}, vol.~51, pp. 3145--3159, 2018.

\bibitem{machado1992structural}
L.~T. Machado, M.~Desbois, and J.-P. Duvel, ``Structural characteristics of deep convective systems over tropical africa and the atlantic ocean,'' \emph{Monthly Weather Review}, vol. 120, no.~3, pp. 392--406, 1992.

\bibitem{machado1998life}
L.~Machado, W.~Rossow, R.~Guedes, and A.~Walker, ``Life cycle variations of mesoscale convective systems over the americas,'' \emph{Monthly Weather Review}, vol. 126, no.~6, pp. 1630--1654, 1998.

\bibitem{kim2017detection}
M.~Kim, J.~Im, H.~Park, S.~Park, M.-I. Lee, and M.-H. Ahn, ``Detection of tropical overshooting cloud tops using himawari-8 imagery,'' \emph{Remote sensing}, vol.~9, no.~7, p. 685, 2017.

\bibitem{lecun2015deep}
Y.~LeCun, Y.~Bengio, and G.~Hinton, ``Deep learning,'' \emph{nature}, vol. 521, no. 7553, pp. 436--444, 2015.

\bibitem{10500381}
C.~Shi, Z.~Su, K.~Zhang, X.~Xie, X.~Zheng, Q.~Lu, and J.~Yang, ``Cloudfu-net: A fine-grained segmentation method for ground-based cloud images based on an improved encoder-decoder structure,'' \emph{IEEE Transactions on Geoscience and Remote Sensing}, pp. 1--1, 2024.

\bibitem{ma2024multilevel}
X.~Ma, X.~Zhang, M.-O. Pun, and M.~Liu, ``A multilevel multimodal fusion transformer for remote sensing semantic segmentation,'' \emph{IEEE Transactions on Geoscience and Remote Sensing}, 2024.

\bibitem{ronneberger2015u}
O.~Ronneberger, P.~Fischer, and T.~Brox, ``U-net: Convolutional networks for biomedical image segmentation,'' in \emph{Medical Image Computing and Computer-Assisted Intervention--MICCAI 2015: 18th International Conference, Munich, Germany, October 5-9, 2015, Proceedings, Part III 18}.\hskip 1em plus 0.5em minus 0.4em\relax Springer, 2015, pp. 234--241.

\bibitem{chen2018encoder}
L.-C. Chen, Y.~Zhu, G.~Papandreou, F.~Schroff, and H.~Adam, ``Encoder-decoder with atrous separable convolution for semantic image segmentation,'' in \emph{Proceedings of the European conference on computer vision (ECCV)}, 2018, pp. 801--818.

\bibitem{liu2021swin}
Z.~Liu, Y.~Lin, Y.~Cao, H.~Hu, Y.~Wei, Z.~Zhang, S.~Lin, and B.~Guo, ``Swin transformer: Hierarchical vision transformer using shifted windows,'' in \emph{Proceedings of the IEEE/CVF international conference on computer vision}, 2021, pp. 10\,012--10\,022.

\bibitem{reichstein2019deep}
M.~Reichstein, G.~Camps-Valls, B.~Stevens, M.~Jung, J.~Denzler, N.~Carvalhais, and f.~Prabhat, ``Deep learning and process understanding for data-driven earth system science,'' \emph{Nature}, vol. 566, no. 7743, pp. 195--204, 2019.

\bibitem{wang2018rgb}
P.~Wang, W.~Li, P.~Ogunbona, J.~Wan, and S.~Escalera, ``Rgb-d-based human motion recognition with deep learning: A survey,'' \emph{Computer vision and image understanding}, vol. 171, pp. 118--139, 2018.

\bibitem{fang2019gstnet}
S.~Fang, Q.~Zhang, G.~Meng, S.~Xiang, and C.~Pan, ``Gstnet: Global spatial-temporal network for traffic flow prediction.'' in \emph{IJCAI}, 2019, pp. 2286--2293.

\bibitem{jenni2020video}
S.~Jenni, G.~Meishvili, and P.~Favaro, ``Video representation learning by recognizing temporal transformations,'' in \emph{European Conference on Computer Vision}.\hskip 1em plus 0.5em minus 0.4em\relax Springer, 2020, pp. 425--442.

\bibitem{hochreiter1997long}
S.~Hochreiter and J.~Schmidhuber, ``Long short-term memory,'' \emph{Neural computation}, vol.~9, no.~8, pp. 1735--1780, 1997.

\bibitem{wang2017predrnn}
Y.~Wang, M.~Long, J.~Wang, Z.~Gao, and P.~S. Yu, ``Predrnn: Recurrent neural networks for predictive learning using spatiotemporal lstms,'' \emph{Advances in neural information processing systems}, vol.~30, 2017.

\bibitem{wang2018predrnn++}
Y.~Wang, Z.~Gao, M.~Long, J.~Wang, and S.~Y. Philip, ``Predrnn++: Towards a resolution of the deep-in-time dilemma in spatiotemporal predictive learning,'' in \emph{International Conference on Machine Learning}.\hskip 1em plus 0.5em minus 0.4em\relax PMLR, 2018, pp. 5123--5132.

\bibitem{bertasius2021space}
G.~Bertasius, H.~Wang, and L.~Torresani, ``Is space-time attention all you need for video understanding?'' in \emph{ICML}, vol.~2, no.~3, 2021, p.~4.

\bibitem{arnab2021vivit}
A.~Arnab, M.~Dehghani, G.~Heigold, C.~Sun, M.~Lu{\v{c}}i{\'c}, and C.~Schmid, ``Vivit: A video vision transformer,'' in \emph{Proceedings of the IEEE/CVF international conference on computer vision}, 2021, pp. 6836--6846.

\bibitem{liu2022video}
Z.~Liu, J.~Ning, Y.~Cao, Y.~Wei, Z.~Zhang, S.~Lin, and H.~Hu, ``Video swin transformer,'' in \emph{Proceedings of the IEEE/CVF conference on computer vision and pattern recognition}, 2022, pp. 3202--3211.

\bibitem{tan2022simvp}
C.~Tan, Z.~Gao, S.~Li, and S.~Z. Li, ``Simvp: Towards simple yet powerful spatiotemporal predictive learning,'' \emph{arXiv preprint arXiv:2211.12509}, 2022.

\bibitem{vaswani2017attention}
A.~Vaswani, N.~Shazeer, N.~Parmar, J.~Uszkoreit, L.~Jones, A.~N. Gomez, {\L}.~Kaiser, and I.~Polosukhin, ``Attention is all you need,'' \emph{Advances in neural information processing systems}, vol.~30, 2017.

\bibitem{long2015fully}
J.~Long, E.~Shelhamer, and T.~Darrell, ``Fully convolutional networks for semantic segmentation,'' in \emph{Proceedings of the IEEE conference on computer vision and pattern recognition}, 2015, pp. 3431--3440.

\bibitem{luo2023trcdnet}
C.~Luo, S.~Feng, Y.~Quan, Y.~Ye, X.~Li, Y.~Xu, B.~Zhang, and Z.~Chen, ``Trcdnet: A transformer network for video cloud detection,'' \emph{IEEE Transactions on Geoscience and Remote Sensing}, 2023.

\bibitem{chen2017rethinking}
L.-C. Chen, G.~Papandreou, F.~Schroff, and H.~Adam, ``Rethinking atrous convolution for semantic image segmentation,'' \emph{arXiv preprint arXiv:1706.05587}, 2017.

\bibitem{wang2021temporal}
H.~Wang, W.~Wang, and J.~Liu, ``Temporal memory attention for video semantic segmentation,'' in \emph{2021 IEEE International Conference on Image Processing (ICIP)}.\hskip 1em plus 0.5em minus 0.4em\relax IEEE, 2021, pp. 2254--2258.

\bibitem{jeppesen2019cloud}
J.~H. Jeppesen, R.~H. Jacobsen, F.~Inceoglu, and T.~S. Toftegaard, ``A cloud detection algorithm for satellite imagery based on deep learning,'' \emph{Remote sensing of environment}, vol. 229, pp. 247--259, 2019.

\bibitem{dronner2018fast}
J.~Dr{\"o}nner, N.~Korfhage, S.~Egli, M.~M{\"u}hling, B.~Thies, J.~Bendix, B.~Freisleben, and B.~Seeger, ``Fast cloud segmentation using convolutional neural networks,'' \emph{Remote Sensing}, vol.~10, no.~11, p. 1782, 2018.

\bibitem{gao2022simvp}
Z.~Gao, C.~Tan, L.~Wu, and S.~Z. Li, ``Simvp: Simpler yet better video prediction,'' in \emph{Proceedings of the IEEE/CVF Conference on Computer Vision and Pattern Recognition}, 2022, pp. 3170--3180.

\end{thebibliography}
%\begin{thebibliography}{1}


%\end{thebibliography}


%\end{IEEEbiography}


%\newpage


\begin{IEEEbiography}
[{\includegraphics[width=1in,height=1.25in,clip,keepaspectratio]{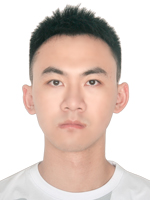}}]{Jiajun Liang}
is currently pursuing the B.S. degree with the School of Computer Science and Technology, Harbin Institute of Technology, Shenzhen, China. His current research interests include computer vision, spatiotemporal data mining, and machine learning.
\end{IEEEbiography}
\begin{IEEEbiography}
[{\includegraphics[width=1in,height=1.25in,clip,keepaspectratio]{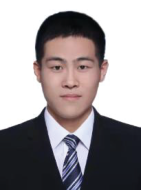}}]{Baoquan Zhang}
is currently an Assistant Professor with the School of Computer Science and Technology, Harbin Institute of Technology, Shenzhen, China. He received the Ph.D. degree in Computer Science from Harbin Institute of Technology, Shenzhen, China, in 2023. His current research interests
include meta learning, few-shot learning, and machine learning.
\end{IEEEbiography}

\begin{IEEEbiography}
[{\includegraphics[width=1in,height=1.25in,clip,keepaspectratio]{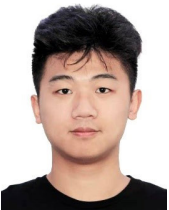}}]{Chuyao Luo}
received the bachelor’s degree in Internet-of-Things engineering from Dalian Maritime University, Dalian, China, in 2017, and the Ph.D. degree in computer science from the Harbin Institute of Technology, Shenzhen, China, in 2022. He is currently a Post-Doctoral Researcher with
the Harbin Institute of Technology. His research interests include data mining, computer vision, time series data prediction, and precipitation nowcasting.
\end{IEEEbiography}

\begin{IEEEbiography}
[{\includegraphics[width=1in,height=1.25in,clip,keepaspectratio]{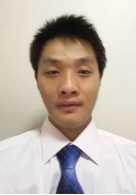}}]{Xutao Li}
	is currently an Professor with the School of Computer Science and Technology, Harbin Institute of Technology, Shenzhen, China. He received the Ph.D. and Master degrees in Computer Science from Harbin Institute of Technology in 2013 and 2009, and the Bachelor from Lanzhou University of Technology in 2007. His research interests include data mining, machine learning, graph mining, and social network analysis, especially tensor-based learning, and mining algorithms.
\end{IEEEbiography}

\begin{IEEEbiography}
[{\includegraphics[width=1in,height=1.25in,clip,keepaspectratio]{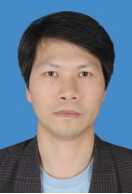}}]{Yunming Ye}
	is currently a Professor with the School of Computer Science and Technology, Harbin Institute of Technology, Shenzhen, China. He received the PhD degree in Computer Science from Shanghai Jiao Tong University, Shanghai, China, in 2004. His research interests include data mining, text mining, and ensemble learning algorithms.
\end{IEEEbiography}

\begin{IEEEbiographynophoto}{Xukai FU}
is the legal representative of Beijing Jinkai New Energy Environmental Technology Co., 2510442827@qq.com, Ltd. His research interests include data mining, text mining, and ensemble learning algorithms.
\end{IEEEbiographynophoto}

%\vfill


\end{document}